\documentclass[acmsmall]{acmart}
\usepackage{amsmath,amsfonts}
\usepackage{algorithmic}
\usepackage{graphicx}
\usepackage{textcomp}
\usepackage{xcolor}
\usepackage{multirow}
\usepackage{times}
\usepackage{soul}
\usepackage{url}
\usepackage{hyperref}
\usepackage{amsmath}
\usepackage{amsthm}
\usepackage{enumitem}
\usepackage{booktabs}
\usepackage{algorithm}
\usepackage{algorithmic}
\usepackage[switch]{lineno}
\usepackage{subfigure}
\usepackage{makecell}
\usepackage{color}
\usepackage{mathrsfs}

\usepackage[]{collab}
\definecolor{ballblue}{rgb}{0.13, 0.67, 0.8}
\definecolor{yellow}{rgb}{1, 0.3, 0.1}
\collabAuthor{jy}{ballblue}{Jinyang Liu}
\collabAuthor{jz}{magenta}{Jiazhen Gu}
\collabAuthor{zb}{teal}{Zhuangbin Chen}

\collabAuthor{ww}{yellow}{Wenwei Gu}

\newcommand{\ie}{{\em i.e.},\xspace}
\newcommand{\eg}{{\em e.g.},\xspace}

\newcommand{\nm}{ISOLATE\xspace}

\AtBeginDocument{
  \providecommand\BibTeX{{
    \normalfont B\kern-0.5em{\scshape i\kern-0.25em b}\kern-0.8em\TeX}}}

\setcopyright{acmcopyright}
\copyrightyear{2024}
\acmYear{2024}
\acmDOI{XXXXXXX.XXXXXXX}

\begin{document}

\title{Identifying Performance Issues in Cloud Service Systems Based on Relational-Temporal Features}

\author{Wenwei Gu}
\orcid{0000-0003-1096-2732}
\affiliation{
  \institution{The Chinese University of Hong Kong}
  \country{Hong Kong SAR}}
\email{wwgu21@cse.cuhk.edu.hk}

\author{Jinyang Liu}
\orcid{0000-0003-0037-1912}
\affiliation{
  \institution{The Chinese University of Hong Kong}
  \country{Hong Kong SAR}}
\email{jyliu@cse.cuhk.edu.hk}

\author{Zhuangbin Chen}
\orcid{0000-0001-5158-6716}
\affiliation{
  \institution{Sun Yat-sen University}
  \country{China}}
\email{chenzhb36@mail.sysu.edu.cn}

\author{Jianping Zhang}
\orcid{0000-0002-8262-9608}
\affiliation{
  \institution{The Chinese University of Hong Kong}
  \country{Hong Kong SAR}}
\email{jpzhang@cse.cuhk.edu.hk}

\author{Yuxin Su}
\orcid{0000-0002-3338-8561}
\affiliation{
  \institution{Sun Yat-sen University}
  \country{China}}
\email{suyx35@mail.sysu.edu.cn}

\author{Jiazhen Gu}
\orcid{}
\authornotemark[0]
\affiliation{
  \institution{The Chinese University of Hong Kong}
  \country{Hong Kong SAR}}
\authornote{Corresponding author.}
\email{jiazhengu@cuhk.edu.hk}

\author{Cong Feng}
\orcid{0009-0000-5556-4004}
\affiliation{
  \institution{Huawei Cloud Computing Technology Co., Ltd}
  \country{China}}
\email{fengcong5@huawei.com}

\author{Zengyin Yang}
\orcid{0000-0001-6307-7310}
\affiliation{
  \institution{Huawei Cloud Computing Technology Co., Ltd}
  \country{China}}
\email{yangzengyin@huawei.com}

\author{Yongqiang Yang}
\orcid{0000-0001-9733-4346}
\affiliation{
  \institution{Huawei Cloud Computing Technology Co., Ltd}
  \country{China}}
\email{yangyongqiang@huawei.com}

\author{Michael R. Lyu}
\orcid{0000-0002-3666-5798}
\affiliation{
  \institution{The Chinese University of Hong Kong}
  \country{Hong Kong SAR}}
\email{lyu@cse.cuhk.edu.hk}

\renewcommand{\shortauthors}{Gu, et al.}

\begin{abstract}
    Cloud systems, typically comprised of various components (\eg microservices), are susceptible to performance issues, which may cause service-level agreement violations and financial losses. Identifying performance issues is thus of paramount importance for cloud vendors. In current practice, crucial metrics, \ie key performance indicators (KPIs), are monitored periodically to provide insight into the operational status of components. Identifying performance issues is often formulated as an anomaly detection problem, which is tackled by analyzing each metric independently. However, this approach overlooks the complex dependencies existing among cloud components. Some graph neural network-based methods take both temporal and relational information into account, however, the correlation violations in the metrics that serve as indicators of underlying performance issues are difficult for them to identify. Furthermore, a large volume of components in a cloud system results in a vast array of noisy metrics. This complexity renders it impractical for engineers to fully comprehend the correlations, making it challenging to identify performance issues accurately. To address these limitations, we propose Identifying Performance Issues based on Relational-Temporal Features (\nm), a learning-based approach that leverages both the relational and temporal features of metrics to identify performance issues. In particular, it adopts a graph neural network with attention to characterizing the relations among metrics and extracts long-term and multi-scale temporal patterns using a GRU and a convolution network, respectively. The learned graph attention weights can be further used to localize the correlation-violated metrics. Moreover, to relieve the impact of noisy data, \nm utilizes a positive unlabeled learning strategy that tags pseudo labels based on a small portion of confirmed negative examples. Extensive evaluation on both public and industrial datasets shows that \nm outperforms all baseline models with 0.945 F1-score and 0.920 Hit rate@3. The ablation study also proves the effectiveness of the relational-temporal features and the PU-learning strategy. Furthermore, we share the success stories of leveraging \nm to identify performance issues in Huawei Cloud, which demonstrates its superiority in practice.
\end{abstract}



\maketitle

\section{Introduction}

Cloud computing has surged in popularity in recent years. Large-scale cloud vendors, \eg Microsoft Azure, Amazon Web Services, and Google Cloud Platform, provide customers with various services over the Internet~\cite{qian2009cloud}. Due to the scale and complexity, performance issues are inevitable~\cite{jiang2009system} in cloud systems. Such performance issues may degrade overall availability and increase service response time, thus causing SLA (Service Level Agreement) violations and substantial economic losses~\cite{amannejad2015detecting, li2022going}. Therefore, identifying performance issues accurately is a critical task during the maintenance of cloud systems~\cite{ibidunmoye2015performance}.

In current practice, cloud vendors typically collect crucial metrics (\ie Key Performance Indicators, KPI), such as CPU utilization and network latency, and then analyze the collected metrics to identify performance issues. Simple as the process might seem, it is a non-trivial task. In particular, a cloud system is typically vast in scale and consists of tremendous software and hardware components (\eg microservices, and virtual machines and servers)~\cite{peiris2014pad, guo2020graph, yu2023logreducer}. Each component may have tens of metrics to be collected in the backend monitoring system, resulting in a large volume of monitoring metrics~\cite{padhy2013big,ward2012semantic}. Since analyzing tremendous data manually is labor-intensive and error-prone~\cite{chen2021graph}, automatic approaches that help on-call engineers identify performance issues are preferred.

In the literature, the performance issue identification problem is generally formulated as an anomaly detection problem based on multivariate metrics~\cite{borghesi2019anomaly, scheinert2021learning, xu2018unsupervised}. Some existing approaches~\cite{zhao2021predicting, shen2020timeseries, su2019robust} detect anomalies based on individual metrics, \ie they only consider the temporal abnormal patterns on individual metrics. However, modern cloud service typically consists of various components (\eg storage, computing, middleware), whose corresponding monitoring metrics have complicated inter-dependencies~\cite{chhetri2016cl,scheinert2021learning, luo2021characterizing}. Take disk I/O operations per second and network throughput as an example: Under normal operating conditions, there is typically a correlation between these two metrics; as disk I/O operations increase due to higher data processing and storage demands, network throughput also increases as more data is being transferred to and from the storage systems. However, if a performance issue arises, this correlation can be violated. For instance, if disk I/O remains high while network throughput suddenly drops significantly, it could indicate a network bottleneck, a network gateway failure, or a network misconfiguration. This deviation from the expected correlation between disk I/O and network throughput can thus be a strong indicator of underlying performance issues in the cloud service.

To alleviate the drawback of overlooking correlations, several graph neural network-based approaches that take the spatial-temporal dependency between multiple metrics~\cite{liu2023practical, zhao2020multivariate, chen2021learning, deng2021graph, han2022learning} are proposed. Though these approaches consider the correlation between metrics, these correlations are typically embedded within the latent feature representation in an implicit manner. In other words, these approaches essentially detect temporal anomalies in these latent representations. In contrast, we propose a correlation violation-aware approach that incorporates the correlation violation explicitly. To effectively detect correlation-violated performance problems, it is essential to account for the correlation violation among metrics as performance anomaly criteria.

However, there are four challenges in identifying performance issues through correlation violations. Firstly, it is hard to automatically and effectively model the complicated correlations among a variety of metrics. A cloud application could comprise tens of microservices. Each microservice may still have tens of metrics, leading to a large number of metrics. Moreover, the correlation between metrics is complicated. Even experienced engineers cannot comprehensively clarify the relations among different metrics. Secondly, even with these correlations extracted, the process of accurately identifying the correlation violations from these intricate inter-metric correlations and then determining the happening of performance issues is a non-trivial task. Thirdly, due to the huge volume of metrics, it is still time-consuming for engineers to investigate the underlying issues simply given a binary (\ie normal and abnormal) result. Automatically pinpointing the anomalous metrics is required. Finally, metric data from large-scale production systems are noisy~\cite{chen2022adaptive}, \ie mild performance issues without obvious anomalies in the metric data. This can blur the distinction between normal and anomalous instances, leading to more false positives and false negatives. Unfortunately, it is infeasible to annotate a huge volume of noisy data manually, while simply neglecting such noises may lead to unsatisfying results.

To address these challenges, we propose Identifying Performance Issues Based on Relational-Temporal Features (\nm), an automated approach to identify performance issues through both capturing the violation of relational features and detecting temporal anomalous features of metrics. To solve the challenge of intricate correlations, \nm utilizes graph neural networks to capture the complicated relational features among metrics explicitly, namely, it employs the graph attention mechanism to characterize the correlations between metrics. To effectively identify the correlation violation between metrics, relational embedding, which is uncoupled from the temporal information of raw metrics, is utilized and fed to downstream anomaly detection components. The correlation-violation metrics are highlighted to provide insights to software reliability engineers like the root cause of the performance issue. Furthermore, to relieve the impact of noisy data, Positive Unlabeled learning (PU learning) is adopted, which finds positive samples and updates the models iteratively. Since PU learning produces both negative and positive samples, while traditional Variational Auto Encoder-based (VAE-based) methods can only take negative examples, \nm employs a novel LC-VAE (Label-Conditional VAE) to distinguish normal and abnormal patterns from the labeled inputs effectively.

To evaluate the performance of \nm, we conduct extensive experiments on a widely-used public dataset and two industrial datasets collected from Huawei Cloud. The experimental results demonstrate that compared with state-of-the-art baselines, \nm achieves the best performance issue identification accuracy with an average F1 score of 0.945. \nm also provides anomalous metrics localization with a Hit rate@3 of 0.920. Moreover, we also conduct ablation studies to validate the effectiveness of our design. A case study with two real cases in Huawei Cloud further shows the practical usefulness of our proposed \nm.

We summarize the main contributions of this work as follows: 

\begin{itemize}[leftmargin=*]
    \item {We propose an end-to-end model that captures the relational violation among monitoring metrics explicitly, offering a more precise way to identify correlation-violated performance anomalies. The violated correlations are utilized to localize the root cause of the performance anomaly. Besides, to alleviate the negative effect of concealed noise in training data, we adopt positive unlabeled learning (PU Learning) on metrics performance anomaly detection and propose the Label-Conditional Variational Autoencoder (LC-VAE) that works seamlessly with PU Learning.}
    \item {We conduct extensive evaluations of \nm on two industrial datasets collected from large-scale online service systems of Huawei Cloud and a publicly available dataset. The results demonstrate that \nm outperforms eighteen state-of-the-art methods.}
    \item {We have successfully deployed \nm into the troubleshooting system of Huawei Cloud. The success stories of our deployment confirm the applicability and effectiveness of our method.}
\end{itemize}

\section{Background}

\label{Background}
In this section, we first introduce the background knowledge about performance issue detection in modern cloud systems. Then, we present an example of an industrial scenario that motivates this work. Finally, we summarize some typical performance issues due to the violation of the correlation of monitoring metrics, which aims to provide a comprehensive understanding of the performance issues and how the correlation of metrics plays a crucial role in detecting and diagnosing performance issues.  

\subsection{Monitoring Metrics in Cloud Service Systems}

\begin{figure}[t]
\centering
\includegraphics[width=.8\linewidth]{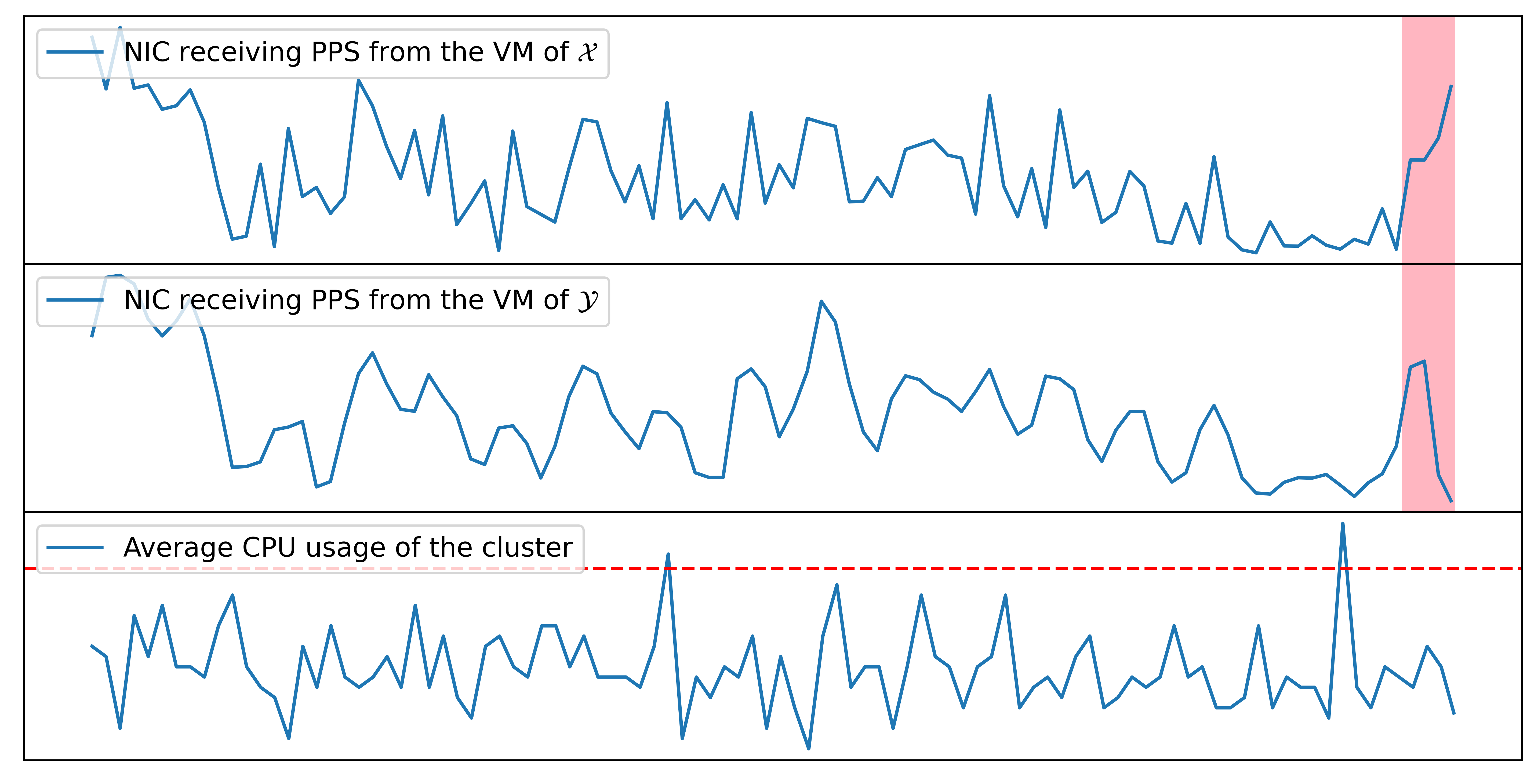}
\vspace{-2mm}
\caption{A real-world example of performance anomaly}
\vspace{-4mm}
\label{Motivating Example}
\end{figure}

In recent years, cloud service systems have gained significant attention due to their ability to provide scalable, on-demand resources and services. Typically, coupled multivariate metrics are collected in run-time to monitor the overall status of the cloud service systems. The collected metrics provide insights into the performance of logical and physical resources within the system, allowing operators to identify and address performance issues before they lead to service disruptions. However, due to the complex inter-dependencies between system components~\cite{ghosh2022fight}, these metrics are often strongly correlated with each other, reflecting the interconnected nature of the system that makes it challenging to isolate and identify specific performance issues. For example, the performance of a virtual machine may be influenced by the workload placed on the underlying physical server, and the performance of a microservice may depend on the performance of other microservices it interacts with. As a result, the metrics collected from different components can exhibit strong correlations with each other, reflecting the complex inter-dependencies within the system. 

\subsection{Performance Issues due to Correlation Violation}
\label{background: correlation}

Performance issues have emerged as a primary concern, potentially undermining the effectiveness of cloud service systems. Some typical factors contributing to these performance issues are resource overload and network latency. Resource overload occurs when the demand for resources exceeds the available capacity, resulting in performance degradation and potential service disruptions~\cite{milani2016load}. Network latency, exacerbated by the distributed nature of cloud systems, can result in reduced application responsiveness, impacting the overall user experience~\cite{bali2013effect}. While these performance issues can be detected by considering single metrics, there are other performance issues that are caused by the violation of correlations between different system metrics, which do not arise from a single metric exceeding a threshold but from a violation in the expected correlation between metrics. Considering the intrinsic correlations between metrics in cloud service systems, accurately capturing and localizing these correlation violation metrics is of great significance to performance issue identification. By localizing anomalous metrics, we gain valuable insights into the specific metrics that deviate from normal behavior (\ie{violation of correlation to other metrics}). Through accurate metric localization, system engineers can swiftly identify the entities that undergo performance anomalies, facilitating prompt troubleshooting and proactive issue resolution.

An industrial case in the online service system of Huawei Cloud is shown in Figure~\ref{Motivating Example}. Three metrics of the elastic load balance (ELB) service are shown in the figure for convenience of presentation. Specifically, the first metric represents the network interface card (NIC) receiving packets per second (PPS) from a virtual machine running microservice $\mathcal{X}$, and the second metric represents the NIC receiving PPS from a virtual machine running microservice $\mathcal{Y}$, which is the downstream microservice that processes the outputs of $\mathcal{X}$. The third metric is the average CPU usage of all virtual machines of the ELB service. We can observe that the variation tendencies between the first and second metrics are somehow consistent during the anomaly-free period. The third metric is used to monitor the resource usage of the service and can raise alerts when there are resource overload issues. However, with the existence of load balance~\cite{ghomi2017load}, a sudden spike in CPU usage will not necessarily lead to a performance issue. Thus, if we trigger alerts based on pre-defined thresholds (as the red dashed line shows), many false alarms will be reported and cause an alert storm that aggravates the burden of engineers~\cite{zhao2020understanding}.

\begin{figure}[t]
\centering
\includegraphics[width=.8\linewidth]{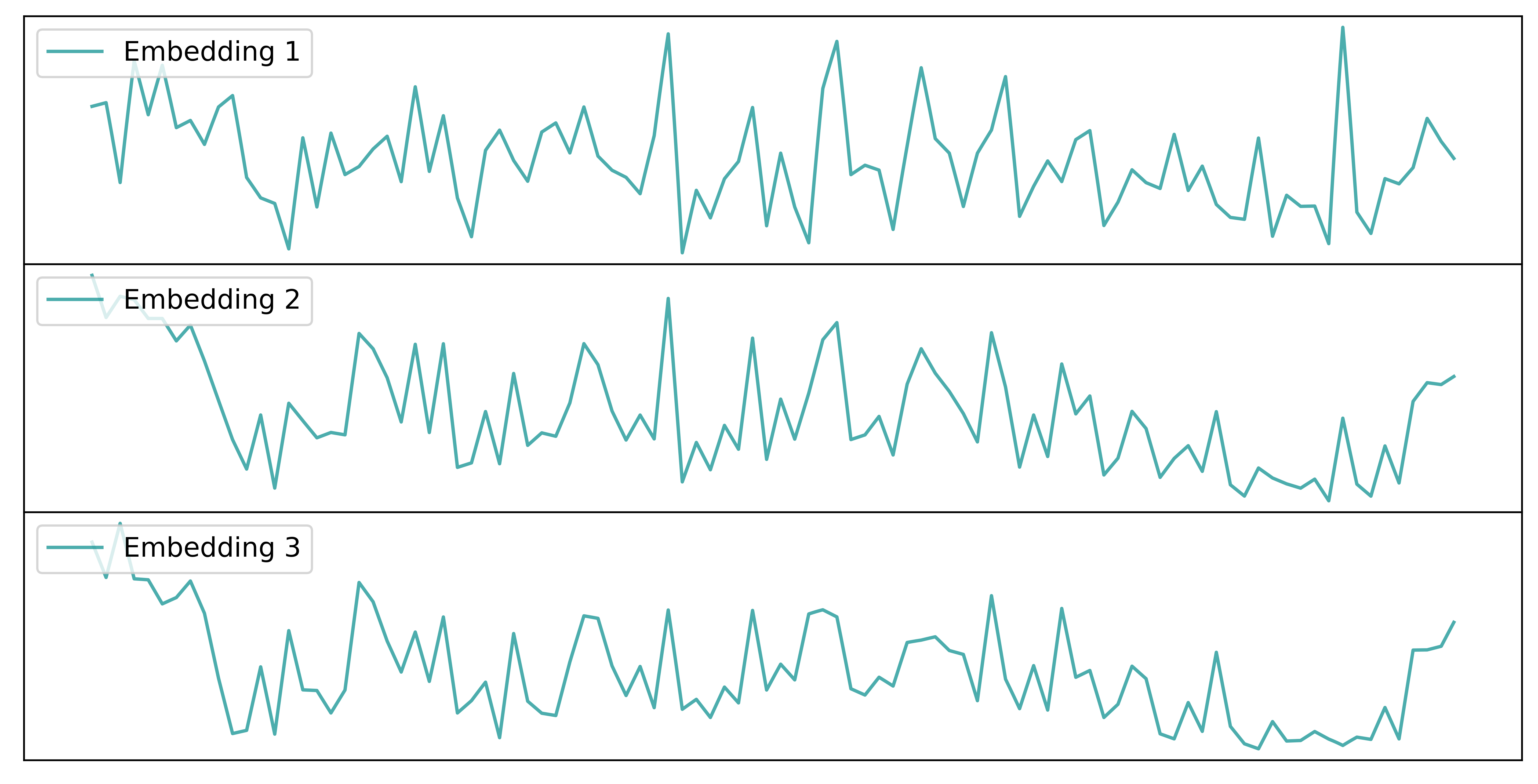}
\vspace{-2mm}
\caption{The latent embedding of graph attention layer}
\vspace{-4mm}
\label{GAT Example}
\end{figure}

The red areas in Figure~\ref{Motivating Example} denote a confirmed network congestion issue by engineers. This issue is caused by a network device failure affecting communication between virtual machines running microservices $\mathcal{X}$ and $\mathcal{Y}$. As a result, the NIC receiving PPS from the virtual machine running $\mathcal{Y}$ suddenly drops even if the NIC receiving PPS from the virtual machine running $\mathcal{X}$ increases. In this case, the correlation between the first and second metrics is critical to rapidly identifying this issue. 

\begin{table}[t]
\centering
\vspace{-4mm}
\caption{A List of Typical Performance Issues Caused by Correlation Violation}
\scalebox{.85}{\begin{tabular}{m{0.18\columnwidth}<{\centering}|m{0.24\columnwidth}<{\centering}|m{0.5\columnwidth}<{\raggedright}}
\hline
Performance Issues & Metrics & Description\\  
\hline
Disk Read/Write Delay  & \makecell{CPU I/0 wait time \\ Disk I/O} & High CPU I/O wait time with low Disk I/O indicates potential disk failure or disk controller problems\\
\hline
Memory Leak & \makecell{Memory slab size \\ Garbage collection time} & High memory slab size with high garbage collection time indicates potential memory leak\\
\hline
API Server Issue & \makecell{API requests \\ CPU usage} & High API requests processed with low CPU usage indicates a potential problem with the API server or an overly efficient request handling mechanism\\
\hline
Network Congestion & \makecell{NIC $A$ receiving PPS \\ NIC $B$ receiving PPS} & High NIC $A$ receiving PPS with low receiving PPS of NIC $B$ or vice versa can indicate potential network congestion\\
\hline
Hardware Interrupt Issue & \makecell{Interrupt requests \\ CPU usage} & High CPU interrupt requests with low CPU usage indicates potential hardware issue or inefficient interrupt handling\\
\hline
Software Interrupt Issue & \makecell{Soft interrupt requests \\ CPU usage} & High CPU soft interrupt request with low CPU usage indicates potential software issue or inefficient interrupt handling\\
\hline
Network Buffering Issue & \makecell{NAT gateway PPS \\ Memory usage} & High NAT gateway PPS with low memory usage indicates potential efficient network buffering mechanism or a network issue\\
\hline
Database Indexing Issue & \makecell{Database query \\ Memory usage} & Slow database query latency with low memory usage indicates potential issues with database indexing or query optimization\\
\hline
\end{tabular}}
\vspace{-4mm}
\label{Performance Issues}
\end{table}

Some of the typical performance issues due to correlation violation are listed in Table \ref{Performance Issues}. Specifically, we take the memory leak as another example to illustrate how the correlation violation reflected on system monitoring metrics can imply performance issues. In cloud service systems, efficient memory management is of great importance and has a direct impact on both the scalability and the operational costs of the cloud environment~\cite{maenhaut2020resource}. A crucial part of the memory management strategy is using slab memory allocation~\cite{jin2019hotspot}, which are pre-allocated extents in persistent memory and containers of free blocks~\cite{dang2022nvalloc} that aid in efficient memory usage. While the garbage collection manages application memory automatically in virtual machines~\cite{degenbaev2018cross}, works to obtain the lifetime of the data objects and then allocates and releases memory space accordingly~\cite{shi2019deca}. Memory leaks occur when a program fails to track and release allocated memory back to the system after it’s no longer needed~\cite{wen2020memlock}. Usually, it is observed that garbage collection time or slab size increases under heavy workloads. An increase in slab size, with a relatively stable garbage collection time, could indicate that the system is effectively managing memory by creating new slabs to handle the increased workload. Similarly, a rise in garbage collection time, with a relatively stable slab size, might suggest that the system reuses existing memory efficiently, resulting in more objects being created and, hence, more garbage to collect. Generally speaking, software systems should efficiently manage memory under heavy workloads, either by creating new memory blocks (indicated by increased slab size) or working harder to clear up unused objects (indicated by increased garbage collection time). Thus, these two cases are not typically considered performance anomalies. However, when both the slab size and garbage collection time consistently increase, even when the workload has reduced, it could suggest a memory leak because memory is being allocated faster than it's being released. It should be noted that a heavy workload can cause these two metrics to increase, hinting at a memory leak when such a leak exists. However, a heavy workload itself will not cause a memory leak, as the root cause of memory leaks lies in programming errors.

It is worth noting that a straightforward way to identify this issue is to build a new metric that combines the first metric and the second metric. However, since a cloud service system typically has a variety of metrics, it is infeasible for engineers to design such combined metrics comprehensively. To alleviate this problem, some GNN-based methods like Graph Attention (GAT) attempt to embed these correlations into latent representations. Nevertheless, we have identified that these methods incorporate correlation information into the latent feature representation implicitly through the combination of all metrics with attention weights. Then the temporal anomalies in these latent representations are detected. The GAT embedding of the example in Figure~\ref{Motivating Example} is shown in Figure~\ref{GAT Example}, where we can observe that the correlation violation between the first and second metrics is even impaired, as the uptrend of the first metric and the downtrend of the second metric are counteracted. Thus, it is crucial to isolate the correlation from the temporal information to detect violation-related performance anomalies.

\section{METHODOLOGY}

In this section, we present \nm, an approach for identifying performance issues based on multivariate metrics in cloud systems. First, we will give a formal definition of the problem. Then, we introduce the overview of \nm and illustrate the design details of our proposed relational-temporal embedding, Label-Conditioned Variational Auto-encoder (LC-VAE), and a novel positive unlabeled strategy. Eventually, we will introduce how to localize the problematic metrics with the correlations obtained in the previous stage.

\subsection{Problem Formulation}

Our objective is to identify performance issues from a variety of monitoring metrics by identifying when the performance anomalies happen and pinpointing the metrics that exhibit temporal or relational anomalies. Specifically, a group of metrics can be seen as a multivariate time series $X\in{R^{N\times{M}}}$, where $N$ denotes the number of observations collected at an equal interval, \ie the length of a time series ~\cite{su2019robust} and $M$ is the number of metrics. $x_t=[x_t^1,x_t^2,...,x_t^M]$ is an $M$-dimensional vector~\cite{hundman2018detecting} that reflects the running status of the system at timestamp $t$. While the $N$-dimensional vector $x^k=[x_1^k,x_2^k,...,x_N^k]$ is the $k^{th}$ metric during the whole monitoring period. 
In addition, a sliding window of historical values $x_{t-c:t}^k=[x_{t-c+1}^k, x_{t-c+2}^k,...,x_t^k]$ is used for modeling the pattern of a current observation, where $c$ is the length of the sliding window.

To determine whether there is an occurrence of performance issues at observation $x_t$, the anomaly score $s_t\in[0,1]$ that represents the degree of being anomalous for each $x_{t-c:t}$ is calculated. Then, the anomaly result can be obtained by comparing the anomaly score against a pre-defined threshold $\theta$. If $s_t>\theta$, the approach will predict the observation $x_t$ as an anomaly. However, it is still unclear to engineers how the anomaly happens. Thus, a kind of anomaly interpretation can be achieved through anomalous metrics localization, {\ie pinpointing a set of metrics $\{x_{k_1},x_{k_2},...x_{k_r}\}$, that is helpful for engineers to find the monitored components that are related to anomaly by the degree of deviation of temporal pattern or break of correlation with other metrics, where $r$ is the number of metrics that are recommended as the anomalous metrics}.

Normalization is performed on each individual metric to unify the range of all metrics and improve the robustness of our model. We normalize the metrics with the maximum and minimum values first:

\begin{equation}
    x^k_{pre} = \frac{{x^k}-min(x^k)}{max(x^k)-min(x^k)}
\end{equation}

Where $max(x^k)$ and $min(x^k)$ represent the maximum and the minimum value of the training set, are computed within the training data and will be used for testing data. For simplicity, we omit the "pre" subscript in the following elaboration.

\subsection{Overview}
\begin{figure*}
\centering
\includegraphics[width=0.9\textwidth]{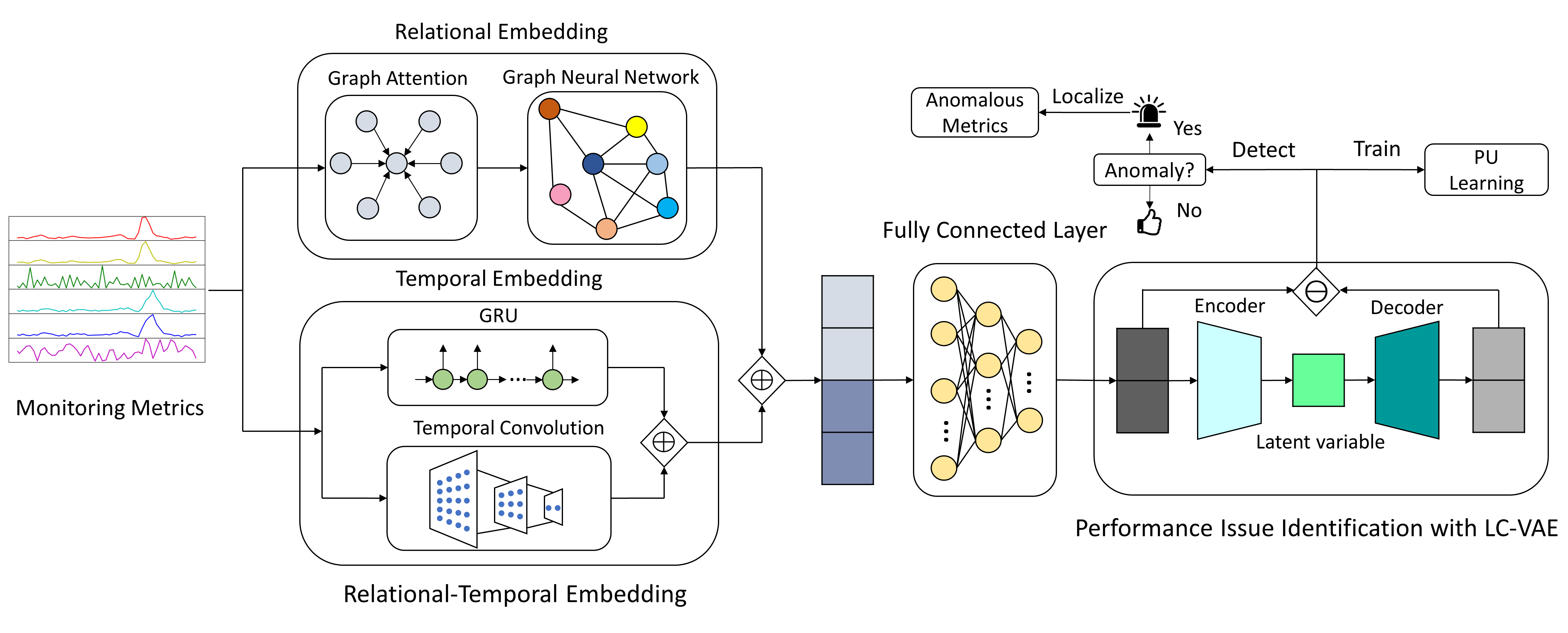}
\vspace{-2pt}
\caption{The Overview of the Proposed Method \nm}
\vspace{-4pt}
\label{ISOLATE}
\end{figure*}

We propose \nm, an automated method that learns correlations among metrics, detects performance anomalies, and locates anomalous metrics. The overview of \nm is shown in Figure~\ref{ISOLATE}, which contains two main parts: the relational-temporal embedding part and the performance issue identification with the LC-VAE part. Specifically, since both abnormal temporal patterns and correlation violations between metrics can indicate performance issues, in the relational-temporal embedding part, \nm captures the relational and temporal patterns from the original metrics (Section~\ref{Method: RT-Embedding}). In particular, due to the lack of information about the correlation among metrics, a complete graph is constructed, then \nm employs graph attention to extract the correlation among metrics. \nm also captures the temporal pattern of each metric through GRU and temporal convolution. In the performance issue identification part, given the inevitable presence of background noise within the data, we propose to use the positive unlabeled learning (PU Learning) strategy during the training phase of \nm (Section~\ref{Method: PU}), which identifies positive samples in unlabeled training data, avoiding the impact of noisy data. As PU learning leads to pseudo-labeled data, a novel label-conditional-VAE (LC-VAE) is adopted to distinguish anomalies from normal patterns (Section~\ref{Method: LC-VAE}). Unlike existing VAE-based methods that solely learn from normal samples, the LC-VAE can learn features from normal and abnormal samples, achieving better performance. Upon the detection of an anomaly, the correlation learned from relational-temporal embedding can aid in localizing the correlation-violating metrics (Section~\ref{Method: localization}). 

\subsection{Relational-Temporal Embedding}
\label{Method: RT-Embedding}

The relational-temporal embedding part takes a group of metrics as inputs. Relational embedding is designed to extract the intrinsic dependencies between metrics and embed the dependencies as a feature vector. Temporal embedding is used to obtain the temporal patterns of metrics as another feature vector because metrics are time series. 

\subsubsection{Relational Embedding}

Specifically, for multivariate metrics with size $M{\times}N$, we can treat each metric $x_i, (i=1,2,...M)$ as a feature vector. The correlations between nodes can be depicted by an adjacency matrix $A\in{R^{M{\times}M}}$. Since we don't have prior knowledge about the correlation between different metrics, we should construct a fully connected graph. We then adopt graph attention networks (GAT)~\cite{brody2021attentive} to learn the correlation between metrics. The attention score is calculated as follows:

\vspace{-3mm}
\begin{align}
    & a_{ij} = \frac{{\rm{exp}}({p^T{\rm{LeakyReLU}}(w\cdot(x_i\oplus{x_j}))})}{\sum_{k=1}^{M}{\rm{exp}}({p^T{\rm{LeakyReLU}}(w\cdot(x_i\oplus{x_k}))})}
\end{align}

The symbol $\oplus$ represents the concatenation operator between two metrics ${x_i}$ and ${x_j}$, $w\in{R^{2N\times{d}}}$ is a matrix of learnable parameters to aggregate the two features. After a nonlinear activation function LeakyReLU~\cite{xu2015empirical}, another learnable vector $p\in{R^{d}}$ is applied. To make the training process more robust and reduce the impact of noise, a threshold $\alpha$ is set to make the adjacency matrix a sparse binary matrix. In other words, the edge between two nodes will be removed if the attention score is below the threshold $\alpha$. Then, a widely used graph convolution layer~\cite{kipf2016semi} is adopted. The formula of graph convolution is shown as follows:

\vspace{-3mm}
\begin{align}
    \hat{h}^{(l+1)} =& \sigma(\widetilde{D}_l^{-\frac{1}{2}}\widetilde{A}_l\widetilde{D}_l^{-\frac{1}{2}}h^{(l)}\Theta_l)
\end{align}

where $\sigma$ is the ReLU activation function and $\widetilde{A}_l=A_l+I$ is the adjacency matrix at the layer $l$, $\widetilde{D}\in{R^{M{\times}M}}$ is the degree matrix of $\widetilde{A}_l$, $h^{(l)}$ is the output representation of the hidden layer $l$ and $\Theta_l\in{R^{M{\times}F}}$ is a learnable parameter. Due to the limitation of space, we only show one layer in Figure {\ref{ISOLATE}}. 

We further apply graph pooling layers between the graph convolution layers to reduce the number of parameters, which can also avoid overfitting. Specifically, as proposed by~\cite{lee2019self}, self-attention~\cite{vaswani2017attention} is utilized to focus more on important features and less on unimportant features. Thus, self-attention scores can be obtained by using another graph convolution. After obtaining the attention scores $Z$, a portion of the nodes and features will be reserved according to the scores. A hyperparameter $k$ refers to the pooling ratio, and the corresponding nodes with the top $\lfloor{kM}\rfloor$ value of $Z$ will be retained. 

Readout layer {\cite{cangea2018towards}} is useful to aggregate all node features and get a summarized representation, which is used to output the relational embedding. The output of the readout layer is as follows:

\vspace{-3mm}
\begin{align}
    r_l = \frac{1}{N_l}\sum_{n=1}^{N_l}{h_n^{(l)}}\oplus\max_{n=1}^{N_l}{h_n^{(l)}}
\end{align}

where $N_l$ denotes the node number of layer $l$, $h_n^{(l)}$ is the $n^{th}$ node feature of $h^{(l)}$ and $\oplus$ is concatenation operator. The outputs of each layer will go through readout layers and will be added up as the output of the relational embedding module because the features of different graphs with different sparsities can be combined.

\subsubsection{Temporal Embedding}

Existing metrics anomaly detection works {\cite{zhao2021predicting, hundman2018detecting}} utilize LSTM to acquire sequential information as Long-term temporal dependency inherently exists in monitoring metrics {\cite{huang2022transferable}}. However, LSTM suffers from the gradient vanishing problem incurred by long-time lags~\cite{fu2016using}. To overcome the drawbacks of LSTM, we apply a GRU to capture the sequential information $t_g$, especially the long-term pattern in the metrics. Then, global average pooling layers are applied on the time series dimension of the output of GRU to get the temporal embedding.

Temporal convolution is useful for capturing the multi-scale temporal information of metrics. Unlike the existing methods that embed metrics with only recurrent neural networks, we also deploy causal convolution implemented by shifting the output of the 1D convolution. To further increase the receptive field of the convolutions, we use dilated convolutions. Dilated convolution is equivalent to filling the convolution kernel with zero padding so as to get a larger convolutional filter. Thus, we adopt dilated causal convolution (DC convolution)~\cite{oord2016wavenet} to extract the temporal embedding of the metrics as it has advantages over the original convolutional operation with a larger receptive field, which is beneficial to our temporal embedding module as it can capture the behaviors of monitoring metrics at multi-scale. A block that consists of DC convolution, batch normalization~\cite{ioffe2015batch}, and activation function (\emph{i.e.} ReLU) is utilized to form the temporal convolution. Eventually, the temporal embedding will be concatenated to the relational embedding and go through a fully connected layer to get the relational-temporal embedding.

\subsection{Performance Issues Identification with LC-VAE}

With the extracted relational-temporal embedding, we then use a novel Label Conditional VAE to detect performance issues. Unlike traditional autoencoder-based methods {\cite{yan2021unsupervised, su2019robust, xu2018unsupervised, zong2018deep, borghesi2019anomaly}} that only take normal samples as input to capture the distribution of metrics, our proposed Label Conditional VAE takes the label obtained from positive unlabeled learning as a part of the input to further help the model differentiate anomalies from normal data because there exist some mild performance issues that are ignored by engineers in training data. 

\subsubsection{Positive Unlabeled Learning}
\label{Method: PU}

Unsupervised methods assume that the training data is anomaly-free. However, there are inevitably some unlabeled anomalies that are ignored by engineers. This assumption can lead to a decline in the overall accuracy as the presence of these hidden anomalies can skew the model's understanding of normal behavior. In our scenario, we only have a small portion of negative data (representing normal, anomaly-free metrics) with high confidence, but we don't have positive data (representing anomalous samples) because finding positive from a large number of unlabeled samples is like looking for a needle in the ocean.

\begin{figure*}[t]
\centering
\includegraphics[width=0.7\columnwidth]{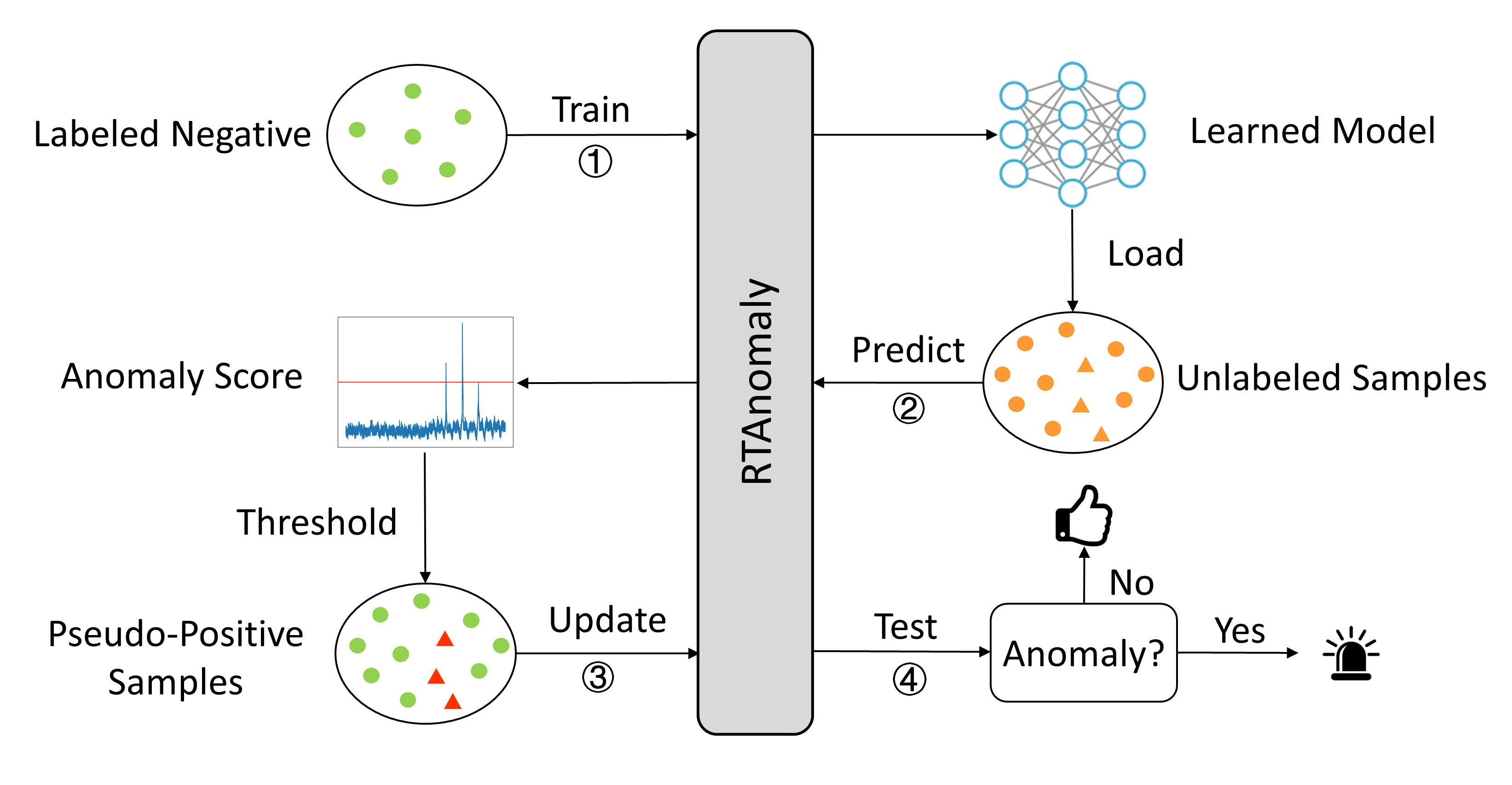}
\vspace{-10pt}
\caption{Positive Unlabeled Learning}
\label{PU Learning}
\vspace{-12pt}
\end{figure*}

To tackle this, \nm tries to find out the anomalous samples using the idea of positive unlabeled learning (PU learning)~\cite{kiryo2017positive}. Specifically, as shown in Figure \ref{PU Learning}, a small amount of negative samples (around 5\%) labeled by engineers is utilized for training the model. These labels are obtained by initially randomly selecting 5\% of the entire training dataset. Engineers then meticulously filter out the noise in these samples and label them as true negatives. This manual labeling process is feasible due to the high confidence in the selected data and can be completed within a few minutes. Consequently, we incorporate human expertise into our method.

With the model trained on these labeled negatives, the remaining unlabeled training data can be predicted. Intuitively, the anomalous samples concealed in the training data are hard to reconstruct with this model and, thus, have a high anomaly score. After obtaining the anomaly score, the samples with an anomaly score that exceeds a pre-defined threshold $\beta$ will be labeled as positive. All the data with pseudo labels are used to update the model. Finally, this updated model will be used to detect performance issues from metrics.

\subsubsection{Label-Conditional VAE}
\label{Method: LC-VAE}
Though the posterior of the distribution $p_\theta(z|y,e)$ is critical for training and prediction of the model, it is hard to obtain. Thus, the variational inference is used to fit a neural network as the approximation posterior $q_\phi(z|y,e)$. Suppose the prior of the latent variable $Z$ is Gaussian distribution $\mathcal{N}(\textbf{0}, \textbf{1})$ and $y$ is the true label of the input sliding windows. Then both posteriors of $e$ and $z$ are chosen to be Gaussian distribution: $p_\theta(e|y,z)=\mathcal{N}(\mu_{\theta}(z), \sigma^2_{\theta}(z))$ and $q_\phi(z|y,e)=\mathcal{N}(\mu_{\phi}(e), \sigma^2_{\phi}(e))$, where $\mu_{\theta}(z)$, $\sigma_{\theta}(z)$ and $\mu_{\phi}(e)$, $\sigma_{\phi}(e)$, are the means and standard deviations of input embedding and latent variable. The input embedding $e$ will be concatenated with the one-hot label vector $y$, and then the latent variable $z$ will be sampled from a posterior $q_\phi(z|y,e)$ at the encoder, which is usually derived by linear layers {\cite{kingma2013auto}}. Eventually, the latent variable will also be concatenated with $y$, and the input embedding will be reconstructed from $p_\theta(e|y,z)$ at the decoder. The reconstructed embedding can be denoted as $\hat{e}$.

In general, the parameters of the LC-VAE can be estimated efficiently with the stochastic gradient variational Bayes (SGVB) algorithm~\cite{rezende2014stochastic}. The evidence-lower bound (ELBO) is a surrogate objective function that can help the estimation. Besides, to better capture the latent pattern of normal sliding windows, we add an additional reconstruction error term. The loss function of the proposed LC-VAE is shown as follows:

\vspace{-3mm}
\begin{equation}
\begin{aligned}
    \mathcal{L}_{LCVAE} =& sgn(0.5-y)*(-KL(q_\phi(z|y,e)\Vert{p_\theta(e|y,z))}\\
    & + \frac{1}{S}\sum_{s=1}^S(\log{ p_{\theta}(e_s|y)} + \lambda\cdot\Vert{e_s-\hat{e_s}}\Vert_2))
\end{aligned}
\end{equation}

Where $S$ is the number of sliding window samples, the first two terms are from evidence-lower bound (ELBO), and the third term is the reconstruction error of the embedding. In this way, we combine the strength of reconstruction and probabilistic estimation together. The coefficient $\lambda>0$ is to trade off the loss terms. As mentioned above, metrics anomaly detection works by learning the normal patterns of sliding windows of metrics; thus, we should minimize the loss function for a coming normal sliding window. However, when there is an anomalous sample input, we should avoid the model to learn the pattern of anomaly by maximizing the loss function. Thus, we denote the loss function with the Signum function to control the sign. In this way, during the detection phase, the anomalies are easy to differentiate by \nm. The reconstruction error $\Vert{e_s-\hat{e_s}}\Vert_2$ will be used as the anomaly score during the detection phase.

\subsection{Correlation Violation Metrics Localization}
\label{Method: localization}
Once an anomaly has been detected in a service system, further analyses will be enacted to determine the possible causes for such a performance anomaly ~\cite{lin2018microscope, shan2019diagnosis, zhong2024automated}. This allows application operators to determine which part of the service this performance issue reported is related to. For monitoring metrics, to further understand the mechanism of performance issues, pinpointing a few metrics that are highly correlated with the root cause is crucial~\cite{yu2023nezha, yu2024changerca}. For example, when we observed that the throughput metrics of two devices were highlighted, it seemed to be a performance issue related to the communication between two devices. While our model highlights the CPU utilization of a service, it is likely to occur due to a lack of computing resources in the run-time environment~\cite{nguyen2011pal}. 

However, existing methods regarding monitoring metrics have not integrated anomaly detection and metric localization, namely root cause localization together in a unified pipeline~\cite{soldani2022anomaly}. In this case, the knowledge during the anomaly detection phase cannot be shared with the localization. We integrate these two closely related tasks into a unified framework to provide more hints to system operators. 

Since there exist correlations between metrics of service in modern online service systems, the learned correlation graph between metrics during the anomaly detection phase is useful for localizing the metrics that reveal the cause of the anomaly ~\cite{wang2021groot, ma2020diagnosing, weng2018root}. Regarding localizing anomalous metrics, correlation information can be more indicative than temporal information. It is straightforward to understand that no matter whether the anomaly is a temporal anomaly that happens on some specific metrics or an anomaly due to the contravention of correlation compared with the anomaly-free stage, the correlation between anomalous metrics and others will undergo drastic changes. However, when some metrics collectively spike, they do not violate the correlations and thus are not anomalous metrics. So, we only utilize the correlations in metrics localization. Particularly based on the above assumption, we can calculate the correlation change as follows:

\begin{equation}
    \Delta{A_i} = \sum_{j\neq{i}}\Vert{A^a_{ij}-A^n_{ij}}\Vert_1
\end{equation}

Where $\Delta{A_i}$ is the variation of correlation between normal and abnormal periods for metric $i$, the $A^n_{ij}$ is computed by averaging $a_{ij}$ on the training period, while the $A^a_{ij}$ is the mean of $a_{ij}$ during the anomaly segment. Eventually, \nm would highlight a few metrics with high $\Delta{A_i}$ and recommend them to engineers to help them get fine-grained information on the performance issue and double-check the devices related to these metrics.

\section{EVALUATION}

To comprehensively evaluate the effectiveness of our proposed approaches \nm, we use both a public dataset and two real-world monitoring metric datasets from the online services of Huawei Cloud Company. Particularly, we aim to answer the following research questions (RQs):

\begin{itemize}[leftmargin=*]
    \item {RQ1: How effective is \nm compared with performance anomaly detection baselines?}
    \item {RQ2: How effective is each component of \nm in performance anomaly identification?}
    \item {RQ3: How effective is \nm in localizing the anomalous metrics?}
    \item {RQ4: How sensitive is \nm to the parameters?}
    \item {RQ5: How efficient is \nm regarding the number of metrics?}
\end{itemize}

\subsection{Datasets}

\begin{table}[t]
\centering
\vspace{-8pt}
\caption{Statistics of Industrial Dataset}
\scalebox{0.85}{\begin{tabular}{ccc}
\toprule
Industrial & Dataset A & Dataset B\\  
\midrule
Services & 21 & 31\\
\midrule
Metrics & 4$\sim$23 & 3$\sim$25\\
\midrule
Train Length & 366,513 & 541,043\\
\midrule
Test Length & 244,356 & 360,716\\
\midrule
Anomaly Ratio & 6.71\% & 5.88\%\\
\bottomrule
\end{tabular}}
\vspace{-8pt}
\label{Datasets}
\end{table}

We conduct experiments on a publicly available dataset. To confirm the practical significance of \nm, we collect two datasets from large-scale online services of Huawei Cloud. The statistics of our industrial datasets are shown in Table {\ref{Datasets}}

\textbf{\emph{Public Dataset}} The public dataset for our experiments is SMD (Server Machine Dataset), which is collected from a large Internet company containing a 5-week-long monitoring metrics of 28 machines {\cite{su2019robust}}. The authors divided the SMD into two subsets of equal size: the first half for the training set and the second half for the testing set. Domain experts labeled anomalies in the SMD testing set based on incident reports.

\textbf{\emph{Industrial Dataset}} To evaluate the effectiveness of \nm in production scenarios, we collect metrics Application CPU Usage, Memory Usage, Interface Throughput, and so on from the online service of the company. We collect metrics with a sampling interval of one minute for more than one week from two regions of the company. The anomalies representing the performance issues of the service are labeled by experienced software engineers with incidents associated with the metrics. The performance issues consist of correlation-violated issues like memory leaks or network congestion listed in Table~\ref{Performance Issues} and non-correlation-violated issues like resource overload. Based on the incident reports, engineers also label the metrics that are correlated to the performance issues. Using these labels, we can also evaluate the accuracy of metrics localization of \nm.

\subsection{Experiment Setting}

\begin{figure*}[t]
\centering
\includegraphics[width=0.5\columnwidth]{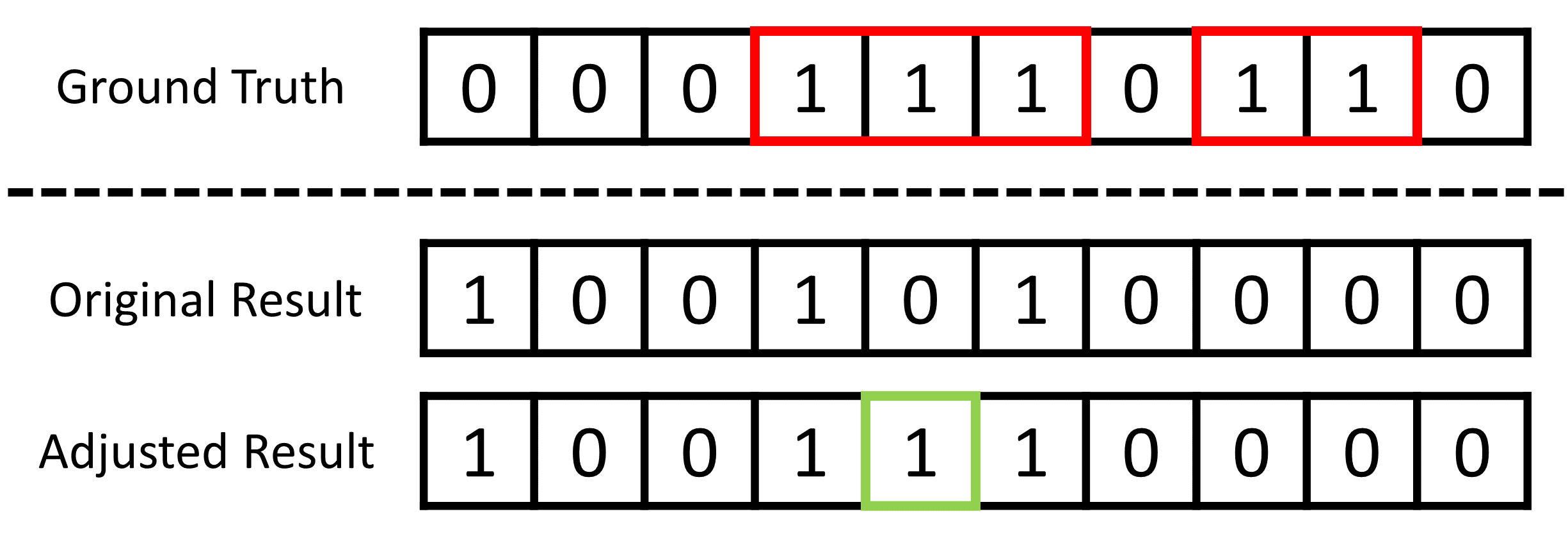}
\vspace{-4pt}
\caption{An Illustration of the Point Adjustment Process in Our Evaluation}
\label{Point Adjustment}
\vspace{-4pt}
\end{figure*}

\subsubsection{Baselines}

The following methods are compared to evaluate the effectiveness of \nm. All the baselines are implemented from the open-sourced codes released by the authors. For both the public dataset SMD and our industrial datasets in Huawei Cloud, we employed a grid search strategy to explore the most suitable parameter configurations. A summary of the baseline methods is shown in Table~\ref{summary}.

\begin{table*}[htbp]
\centering
\caption{A Summary of the Baseline Methods}
\scalebox{0.85}{\begin{tabular}{c|c|c}
\toprule
Method Category & Supervision Type & Reference Baseline Methods\\
\midrule
\multirow{1}{*}{Signal Processing-based} & \multirow{1}{*}{Unsupervised} & JumpStarter~\cite{ma2021jump}\\
\midrule
\multirow{1}{*}{Machine Learning-based} & \multirow{1}{*}{Unsupervised} & LOF~\cite{breunig2000lof}, IForest~\cite{liu2008isolation}, OCSVM~\cite{scholkopf2001estimating}\\
\midrule
\multirow{1}{*}{LSTM-based} & \multirow{1}{*}{Unsupervised} & LSTM~\cite{hundman2018detecting, zhao2021predicting}, LSTM-VAE~\cite{park2018multimodal}, THOC~\cite{shen2020timeseries}\\
\midrule
\multirow{3}{*}{Autoencoder-based} & \multirow{2}{*}{Unsupervised} & DAGMM~\cite{zong2018deep}, OmniAnomaly~\cite{su2019robust}, ACVAE~\cite{zhang2024acvae}\\ 
& & TranAD~\cite{tuli2022tranad}, MTSAD~\cite{wang2022active}, MSCRED~\cite{zhang2019deep}\\
\cmidrule{2-3}
& \multirow{1}{*}{Semi-supervised} & SLA-VAE~\cite{huang2022semi}\\
\midrule
\multirow{1}{*}{Graph Neural Network-based} & \multirow{1}{*}{Unsupervised} & MTAD-GAT~\cite{zhao2020multivariate}, GDN~\cite{deng2021graph}, GTA~\cite{chen2021learning}, FuSAGNet~\cite{han2022learning}\\
\bottomrule
\end{tabular}}
\vspace{0mm}
\label{summary}
\end{table*}

\begin{itemize}[leftmargin=*]
    \item {\textit{OCSVM}~\cite{scholkopf2001estimating}. OCSVM is a clustering-based anomaly detection method that learns the boundary for the normal data points and identifies the data outside the border as anomalies. The input in this baseline is the observation of time series at each timestamp and the output is the anomalous timestamps.}
    \item {\textit{IForest}~\cite{liu2008isolation}. Isolation Forest ensembles a number of isolation trees and recursively partitions the feature space to detect anomalies. The samples with awfully shorter heights are likely to be anomalies. In this baseline, the input is the observation at each timestamp and the output is the anomalous timestamps.}
    \item {\textit{LOF}~\cite{breunig2000lof}. Local Outlier Factor (LOF) is based on density estimation that calculates the local density deviation of a given sample with respect to its neighbors. The anomalies have a substantially lower density than their neighbors. The input is the observation at each timestamp and the output is the anomalous timestamps.}
    \item {\textit{DAGMM}~\cite{zong2018deep}. DAGMM is a model that utilizes an autoencoder to generate a low-dimensional representation and a Gaussian Mixture Model to go through a probabilistic estimation to obtain the anomaly score.}
    \item {\textit{LSTM}~\cite{hundman2018detecting, zhao2021predicting}. LSTM neural network captures the normal behaviors of metrics by forecasting the next values of metrics based on historical observations. Anomalies will be reported if the differences between predicted values and real values exceed a pre-defined threshold}
    \item {\textit{LSTM-VAE}~\cite{park2018multimodal}. LSTM-VAE detects anomalies by integrating LSTM and VAE. It projects observations at each timestamp into a latent space and then estimates the distribution of it using VAE.}
    \item {\textit{OmniAnomaly}~\cite{su2019robust}. OmniAnomaly is a model that captures the normal patterns by learning robust representations of metrics with stochastic Recurrent Neural Network (RNN) and planar normalizing flow based on the reconstruction error.}
    \item {\textit{THOC}~\cite{shen2020timeseries}. THOC is a model that captures the multi-scale temporal features from dilated recurrent layers by a hierarchical clustering mechanism and detects the anomalies by the multi-layer distances.}
    \item {\textit{MTSAD}~\cite{wang2022active}. MTSAD is a deep unsupervised anomaly detection model that incorporates the strength of active learning, including three feedback strategies, namely denominator penalty, negative penalty, and metric learning.}
    \item {\textit{MSCRED}~\cite{zhang2019deep}. MSCRED utilizes convolutional LSTM layers to capture the temporal information and embed the inter-metric information through a signature matrix. It detects anomalies with reconstruction errors of the input metric.}
    \item {\textit{MTAD-GAT}~\cite{zhao2020multivariate}. MTAD-GAT is a graph attention-based model that captures both feature and temporal correlations. It passes these correlations to a Gated-Recurrent-Unit (GRU) network for reconstruction and forecast.}
    \item {\textit{GDN}~\cite{deng2021graph}. GDN learns the graph of relationships between metrics through graph structure learning. It then uses attention-based forecasting and deviation scoring to output anomaly scores.}
    \item {\textit{GTA}~\cite{chen2021learning}. GTA is a transformer-based model that employs graph structure learning to learn the relationship among multiple IoT time series, and the Transformer for temporal modeling and the reconstruction error for anomaly detection.}
    \item {\textit{TranAD}~\cite{tuli2022tranad}. TranAD is a transformer-based model that detects anomalies with reconstruction error by incorporating attention mechanisms and adopting adversarial training.}
    \item {\textit{SLA-VAE}~\cite{huang2022semi}. SLA-VAE is a semi-supervised VAE-based anomaly detection model for online service systems, which employs active learning to update the model.}
    \item {\textit{ACVAE}~\cite{zhang2024acvae}. ACVAE is a self-adversarial variational autoencoder combined with a contrast learning mechanism that allows the encoder to obtain more training samples.}
    \item {\textit{FuSAGNet}~\cite{han2022learning}. FuSAGNet jointly optimizes reconstruction using a sparse autoencoder and forecasting using a graph neural network. It captures the interdependencies between time series in sensors.}
    \item {\textit{JumpStarter}~\cite{ma2021jump}. JumpStarter is a compressed sensing-based method combined with shape-based clustering and an outlier-resistant sampling algorithm. This combination ensures a shorter initialization time.}
\end{itemize}

\subsubsection{Evaluation Metrics}

The anomaly detection problem is modeled as a binary classification problem, so the widely-used binary classification measurements can be applied to evaluate the performance of models. We employ Precision: $PC=\frac{TP}{TP+FP}$, Recall: $RC=\frac{TP}{TP+FN}$, F1 score: $F1=2\cdot\frac{PC\cdot{RC}}{PC+RC}$. Specifically, $TP$ is the number of abnormal samples that the model correctly discovered; $FP$ is the number of normal samples that are incorrectly classified as anomalies; $FN$ is the number of anomalous samples that failed to be detected by the model. F1 score is the harmonic mean of the precision and recall, which symmetrically represents both precision and recall in one metric. Following~\cite{audibert2020usad}, we get the anomaly threshold by grid search for all baselines and \nm to evaluate the performance.

In real-world applications, anomalies will last for a while, leading to consecutive anomalies in the monitoring metrics. Therefore, it is acceptable for the model to trigger an alert for any point in a contiguous anomaly segment if the delay is within the acceptable range. Thus, we adopt the evaluation strategy following {\cite{ren2019time, su2019robust, tuli2022tranad}} that marks the whole segment of continuous anomalies as an anomaly. In other words, we consider the model to correctly predict an anomalous segment if at least one timestamp is successfully predicted as an anomaly. An example of the point adjustment strategy is shown in Figure {\ref{Point Adjustment}}. The first anomaly segment is treated as correctly predicted, so the second point of the segment is adjusted to correctly predicted as anomalous. 

\subsubsection{Implementations} 

We run all the experiments on a Linux server with Intel Xeon Gold 6140 CPU @ 2.30GHZ and Tesla V100 PCIe GPU. The proposed model is implemented under the PyTorch framework and runs on the GPU. The hidden sizes of the GAT layer, GRU layer, and temporal convolution layers are 256, 128, and 128. The coefficient of loss function $\lambda$ is 0.5. The dimension of the latent variable in LC-VAE is 10. The threshold of positive learning is 0.9. We train \nm with the Adam optimizer~\cite{kingma2014adam} with a learning rate of 0.001, a batch size of 128, and an epoch number of 50. We have released the artifacts and data for future research purposes on {\href{https://github.com/WenweiGu/ISOLATE}{https://github.com/WenweiGu/ISOLATE}}.

\subsection{Experimental Results}

\subsubsection{\textbf{RQ1} The effectiveness of \nm}

\begin{table*}[t]
\centering
\caption{Experimental Results of Different Anomaly Detection Methods.}
\scalebox{0.85}{\begin{tabular}{c|ccc|ccc|ccc}
\toprule
\multirow{2}{*}{Methods} & \multicolumn{3}{c|}{SMD} & \multicolumn{3}{c|}{Dataset A} & \multicolumn{3}{c}{Dataset B}\\  
 & Precision & Recall & F1 & Precision & Recall & F1 & Precision & Recall & F1\\ 
\midrule
OCSVM & 0.4434 & 0.7672 & 0.5619 & 0.8639 & 0.5491 & 0.6446 & 0.8979 & 0.5955 & 0.6547\\
IForest & 0.4231 & 0.7329 & 0.5364 & 0.9375 & 0.5782 & 0.6726 & 0.9443 & 0.6713 &
0.7431\\
LOF & 0.5634 & 0.3986 & 0.4668 & 0.9218 & 0.5744 & 0.6693 & 0.9583	& 0.4302 & 0.6160\\
DAGMM & 0.5951 & 0.8782 & 0.7094 & 0.7514 & 0.8108 & 0.7792 & 0.8112 & 0.9073 & 0.8421\\
LSTM & 0.7855 & 0.8528 & 0.8178 & 0.9177 & 0.8107 & 0.8366 & 0.9166 & 0.6994 & 0.7665\\
LSTM-VAE & 0.8698 & 0.7879 & 0.8083 & 0.8753 & 0.7443 & 0.7936 & 0.7969 & 0.7714 & 0.7839 \\
OmniAnomaly & 0.8368 & 0.8682 & 0.8522 & 0.8769 & 0.9084 & 0.8892 & 0.9437 & 0.7985 & 0.8607\\
THOC & 0.7976 & 0.9095 & 0.8499 & 0.9502 & 0.8022 & 0.8347 & 0.9613 & 0.8400 & 0.8866\\
MTSAD & 0.8745 & 0.9395 & 0.9042 & 0.8719 & 0.9171 & 0.8950 & 0.9377 & 0.8765 & 0.9026\\
MSCRED & 0.7876 & 0.9374 & 0.8434 & 0.8928 & 0.8451 & 0.8570 & 0.9554 & 0.8522 & 0.8838 \\ 
MTAD-GAT & 0.8210 & 0.9215 & 0.8683 & 0.8519 & 0.9019 & 0.8682 & 0.9329 & 0.8769 & 0.9012\\
GDN & 0.7670 & 0.9362 & 0.8312 & 0.8831 & 0.9109 & 0.9018 & 0.9614 & 0.8662 & 0.8994\\
GTA & 0.8768 & 0.8987 & 0.8822 & 0.8671 & 0.9098 & 0.8784 & 0.9431 & 0.8791 & 0.8554\\
TranAD & 0.8882 & 0.9023 & 0.8953 & 0.9275 & 0.8703 & 0.8867 & 0.8816 & 0.8929 & 0.8874\\
SLA-VAE & 0.8672 & 0.9371 & 0.9019 & 0.9523 & 0.8605 & 0.8975 & 0.9361 & 0.8859 & 0.8937\\
ACVAE & 0.8779 & 0.8384 & 0.8612 & 0.8988 & 0.8405 & 0.8645 & 0.9151 & 0.8432 & 0.8563\\
FuSAGNet & 0.8319 & 0.9473 & 0.8736 & 0.9011 & 0.8736 & 0.8790 & 0.9038 & 0.8941 & 0.8978\\
JumpStarter & 0.8913 & 0.9174 & 0.9063 & 0.8427 & 0.7959 & 0.8295 & 0.8697 & 0.7869 & 0.8435\\
\midrule
\textbf{\nm} & \textbf{0.8998} & \textbf{0.9745} & \textbf{0.9346} & \textbf{0.9871} & \textbf{0.9435} & \textbf{0.9497} & \textbf{0.9759} & \textbf{0.9367} & \textbf{0.9514}\\
\bottomrule
\end{tabular}}
\vspace{0mm}
\label{RQ1}
\end{table*}

To answer this research question, we compare the performance of \nm with other state-of-the-art baselines on a public dataset and two industrial datasets. First, We train \nm on a small portion of negative samples. Then, \nm will assign pseudo labels to the remaining training data according to the anomaly score and threshold $\beta$. During the test phase, the anomaly score for each timestamp will be computed. The anomaly threshold will be searched following~\cite{audibert2020usad} to produce the prediction result.

The results are shown in Table {\ref{RQ1}}, where the best F1 scores are marked with boldface. We can see the average F1 score of \nm outperforms all baseline methods in three datasets. The experimental results are shown in Table 2. In Dataset B, the improvement achieved by \nm is more significant as the metrics correlations between metrics in Dataset B are more complicated and the ratio of performance anomalies caused by correlation violation is higher. Generally speaking, \nm’s good performance can be attributed to two reasons: Firstly, the utilization of relational-temporal embedding, as the anomalies can be caused by the violation of correlation, which can hardly be detected by finding the spikes on a single metric, it can also help facilitate localization of the anomalous metrics. Thus, there is a significant improvement in the recall of \nm compared to other baselines. Secondly, \nm effectively learns potential anomalous samples from the training data, thereby mitigating the risk of overfitting anomalous patterns during the training process. Consequently, \nm achieves remarkable precision, ranking among the best when compared to other baseline methods.

\begin{table*}[t]
\centering
\vspace{-2pt}
\caption{Experimental Results of the Ablation Study}
\scalebox{0.85}{\begin{tabular}{c|ccc|ccc|ccc}
\toprule
\multirow{2}{*}{Methods} & \multicolumn{3}{c|}{SMD} & \multicolumn{3}{c|}{Dataset A} & \multicolumn{3}{c}{Dataset B}\\  
 & Precision & Recall & F1 & Precision & Recall & F1 & Precision & Recall & F1\\ 
\midrule
\nm \textit{w/o} $\mathcal{RT}$ & 0.8512 & 0.9079 & 0.8723 & 0.9791 & 0.8285 & 0.8738 & 0.9465 & 0.8879 & 0.9035\\
\nm \textit{w/o} $\mathcal{PU}$ & 0.8730 & 0.9407 & 0.9062 & \textbf{0.9912} & 0.8405 & 0.8869 & 0.9396 & 0.9189 & 0.9231\\
\textbf{\nm} & \textbf{0.8998} & \textbf{0.9745} & \textbf{0.9346} & 0.9871 & \textbf{0.9235} & \textbf{0.9497} & \textbf{0.9759} & \textbf{0.9367} & \textbf{0.9514}\\
\bottomrule
\end{tabular}}
\vspace{-2pt}
\label{RQ2}
\end{table*}

Typically, the baseline methods have higher precision than recall because there are some anomalies that are not very apparent. We can observe that machine learning-based methods, namely OCSVM, IForest, and LOF have relatively low performance compared with other baseline models since these methods learn the metrics pattern at each timestamp independently without considering the temporal dependency. While LSTM-based methods and autoencoder-based methods, including LSTM, DAGMM, and LSTM-VAE, perform notably better than OCSVM, IForest, and LOF and achieve 0.7094$\sim$0.8421 F1 score because these models take the historical observation window of the data, that helps to retain valuable historical temporal pattern. Among the baselines, we can find OmniAnomaly, THOC, MTSAD, MSCRED, TranAD, ACVAE, and SLA-VAE can achieve relatively better performance (especially MTSAD can achieve an F1 score like 0.9042) because these methods introduce some mechanisms to extract temporal information of metrics and ensure robust anomaly detection. It should be noted that SLA-VAE is a semi-supervised VAE-based method and it employs active learning to update the model. However, our proposed method outperforms SLA-VAE, suggesting that the improved performance of our model is not solely due to the semi-supervised mechanism, \ie PU learning, but also the design of combining correlational violation detection with temporal information. Specifically, OmniAnomaly models the metrics through stochastic variables and uses reconstruction probabilities to determine anomalies. As for MSCRED, the temporal information is also captured through convolutional LSTM. Meanwhile, in MTSAD, active learning has been incorporated into a variational autoencoder to ensure capability against noise. SLA-VAE is also a variational autoencoder-based architecture that employs both active learning and semi-supervised learning. Adversarial training is incorporated in the VAE-based model ACVAE and the transformer-based model TranAD, which assures their robustness. In THOC, the complex nonlinear temporal dynamics of the system’s normal behavior are captured. Though extracting temporal information well, the limitation of these approaches lies in not taking the correlation features into consideration, which is essential to successfully detecting anomalies from multivariate metrics. Graph Neural Network (GNN)-based approaches offer some mitigation to this issue through jointly extracting the relational and temporal information of raw metrics, \eg MTAD-GAT, GDN, GTA, and FuSAGNet. However, without explicitly capturing the correlation violation, these approaches still achieve suboptimal performance compared with \nm.

\subsubsection{\textbf{RQ2} The effectiveness of components in \nm}

To answer this research question, we conducted an extensive ablation study on \nm. Particularly, we derive two baseline models based on removing the relational-temporal embedding and positive unlabeled learning parts of \nm to investigate the contribution of these two components.

\begin{itemize}[leftmargin=*]
    \item {\textit{\nm w/o $\mathcal{RT}$} This baseline is a variant of \nm that removes relational-temporal embedding that captures both information of metrics. Instead, only an LSTM layer is utilized in this baseline to embed the temporal information of the raw metrics, which will be further fed into the CVAE.}
    \item {\textit{\nm w/o $\mathcal{PU}$} This baseline removes the positive unlabeled learning that finds anomalous samples from a large number of normal samples. All the training data are considered normal in this baseline.}
\end{itemize}

Table {\ref{RQ2}} shows the experimental results of \nm and its variants. Overall, relational-temporal embedding and positive unlabeled learning help to improve the effectiveness of \nm as it performs the best, while the degree of contribution of relational temporal embedding is larger. We attribute this to the good capability of relational temporal embedding in extracting both the temporal information of each metric and correlational information between metrics. When an anomaly happens due to a breach of relationship during the anomaly-free period, it can be easily identified by \nm, while other methods have difficulty identifying it as they are more effective on temporal outliers. We observe that even without PU learning, our model is not worse than all baselines, which further demonstrates the effectiveness of explicitly extracting correlation violation.

The variant without relational-temporal embedding is similar to LSTM-VAE, which employs LSTM to extract the sequential information and VAE to differentiate the anomaly from normal. However, due to the design of positive unlabeled learning and conditional VAE, the variant can identify the noise of training data and label them as positive. Thus, the performances on three datasets of \nm without relational-temporal embedding in terms of F1 score are improved compared to LSTM-VAE, respectively. We believe that in some scenarios with a higher ratio of noise, positive unlabeled learning would play a greater role in improving performance.

\begin{table}[t]
\centering
\vspace{-6pt}
\caption{Performance on Metrics Localization} 
\scalebox{0.85}{\begin{tabular}{c|cc|cc}
\toprule
\multirow{2}{*}{Methods} & \multicolumn{2}{c|}{Dataset A} & \multicolumn{2}{c}{Dataset B}\\  
 & Hit@1 & Hit@3 & Hit@1 & Hit@3\\ 
\midrule
DAGMM & 0.5714 & 0.6667 & 0.5806 & 0.7419\\
LSTM-VAE & 0.6190 & 0.7142 & 0.6451 & 0.7741\\
\nm-Cor & 0.7419 & 0.8387 & 0.7142 & 0.8571\\
\textbf{\nm} & \textbf{0.8095} & \textbf{0.9048} & \textbf{0.8387} & \textbf{0.9354}\\
\bottomrule
\end{tabular}}
\vspace{-8pt}
\label{RQ3}
\end{table}

\subsubsection{\textbf{RQ3} The effectiveness of \nm in localizing the anomalous metrics}

To further demonstrate the capability of \nm, experiments on localizing the metrics are conducted. Specifically, we localize the metrics by following four methods:

\begin{itemize}[leftmargin=*]
    \item {\textit{DAGMM.} This baseline uses the anomaly score of each metric output by DAGMM as the degree of the metric being anomalous. The metrics with the highest scores will be highlighted as anomalous metrics.}
    \item {\textit{LSTM-VAE.} This baseline uses the anomaly score output by LSTM-VAE as the degree of being anomalous for the metrics.}
    \item {\textit{\nm-Cor.} This baseline sums up the correlation of a specific metric between other metrics as the degree of being the root cause. Metrics with a high correlation with other metrics would be reported as anomalous metrics.}
    \item {\textit{\nm.} Different from \nm-Cor, it uses the discrepancy between the anomaly-free period and anomaly period since the correlation between metrics may undergo drastic changes compared to the normal period when an anomaly happens.}
\end{itemize}

As mentioned in Section~\ref{Background}, a metric that shows very abnormal behaviors compared to the normal period is not necessarily the most anomalous metric because it can have a small influence on other metrics and will self-heal. Using the correlation score itself can also cause inaccurate results because some metrics show consistently high correlations with others, no matter whether it is during an anomaly period or not. Indeed, the correlation difference between normal and abnormal time is a stronger indicator of anomalous metrics because when an anomaly happens, the correlation would be obeyed and cause a huge correlation change.

Table \ref{RQ3} presents the comparison of four methods, and we can observe that metrics localization using \nm outperforms the other three baselines with a Hit@1 of 80.95\%, 83.87\%, and Hit@3 of 90.48\%, 93.54\% on two industrial datasets. Thus, our method can provide accurate hints for engineers on which part of the service system they can remedy to ensure the system's reliability. Compared to using the anomaly score of DAGMM and LSTM, using the correlation score to localize metrics (\nm-Cor) is more effective as the anomaly scores of these two methods are computed on a single metric and are not aware of other metrics, while correlation considers the global information of metrics. Usually, the metrics that have a higher correlation with other metrics seem to play a greater influence on the service system and are more likely to be anomalous metrics when an anomaly happens.

\begin{figure*}
  \centering
  \subfigure[Sensitivity of $\alpha$ on Dataset A]{
  \label{Sensitivity 1}
  \includegraphics[width=0.23\columnwidth]{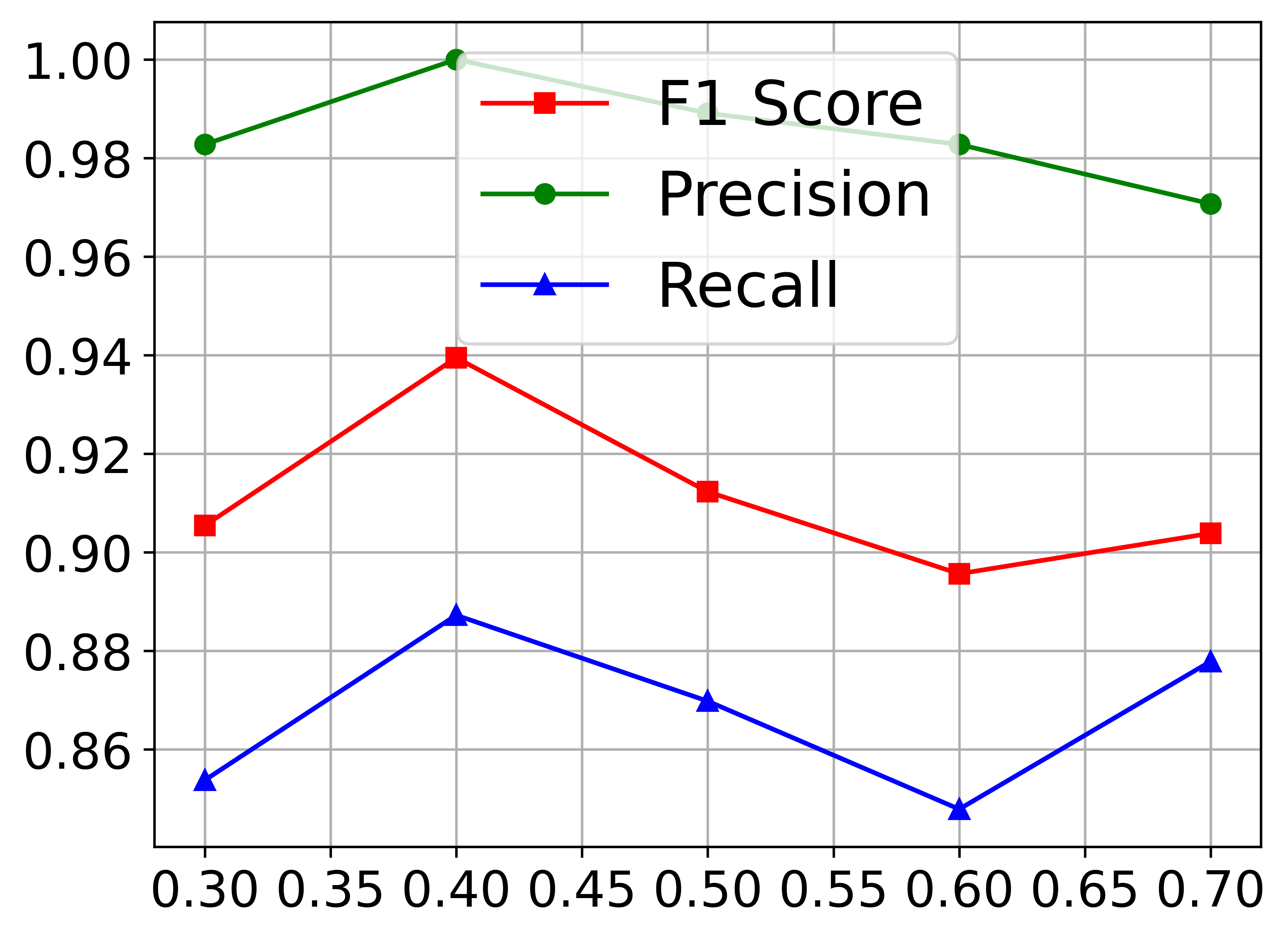}
  }
  \subfigure[Sensitivity of $\alpha$ on Dataset B]{
  \label{Sensitivity 2}
  \includegraphics[width=0.23\columnwidth]{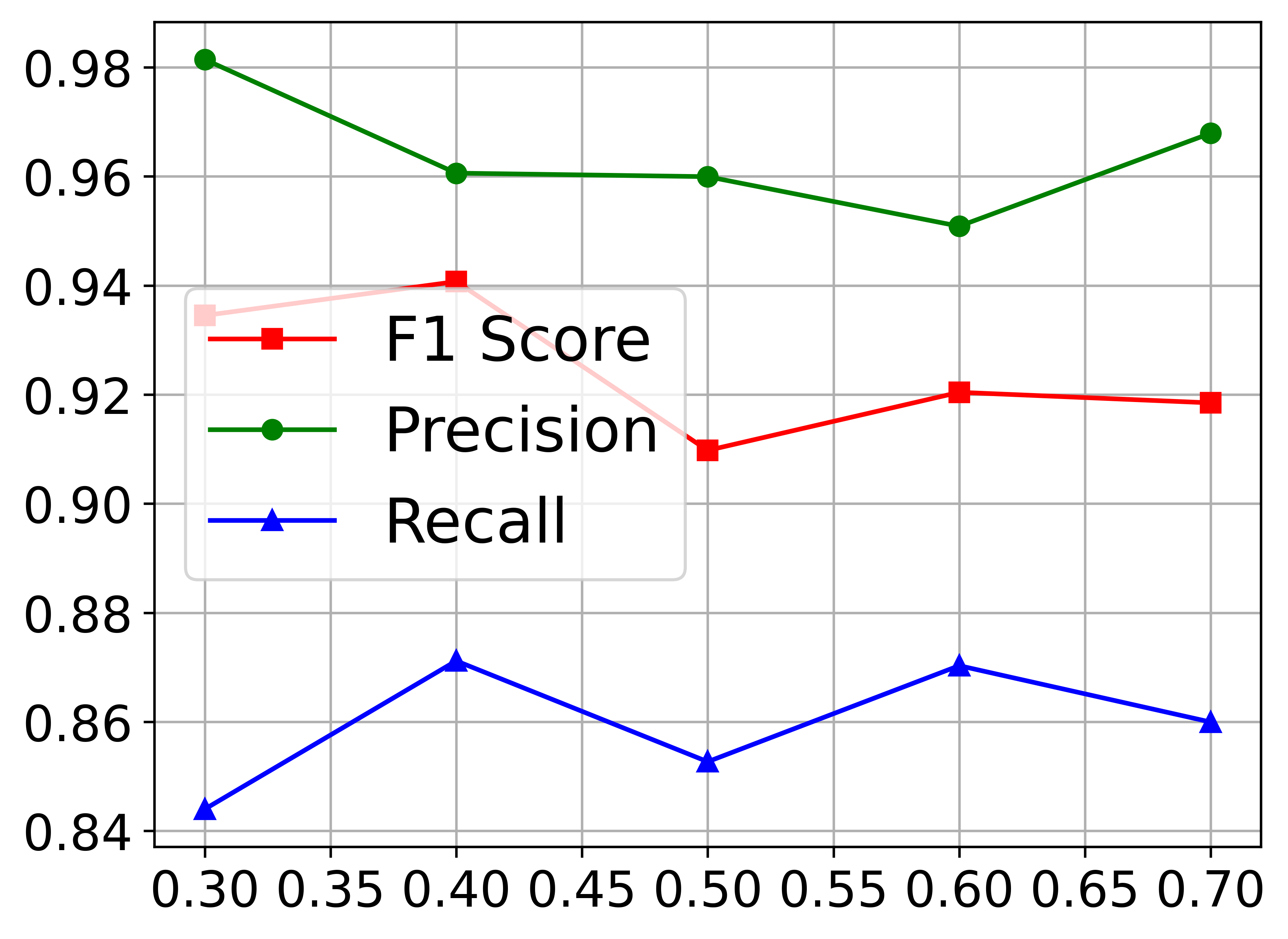}
  }
  \subfigure[Sensitivity of $\beta$ on Dataset A]{
  \label{Sensitivity 3}
  \includegraphics[width=0.23\columnwidth]{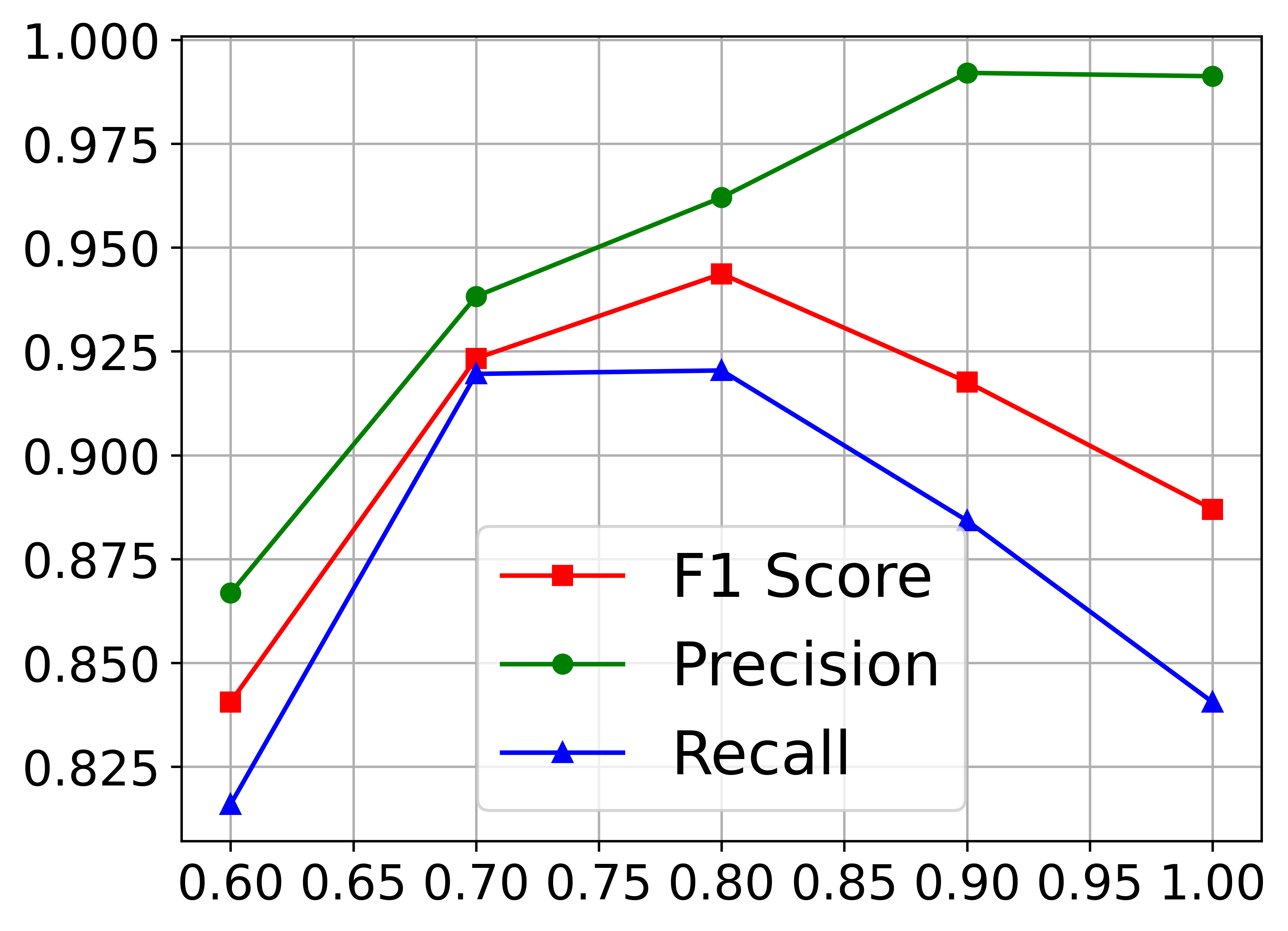}
  }
  \subfigure[Sensitivity of $\beta$ on Dataset B]{
  \label{Sensitivity 4}
  \includegraphics[width=0.23\columnwidth]{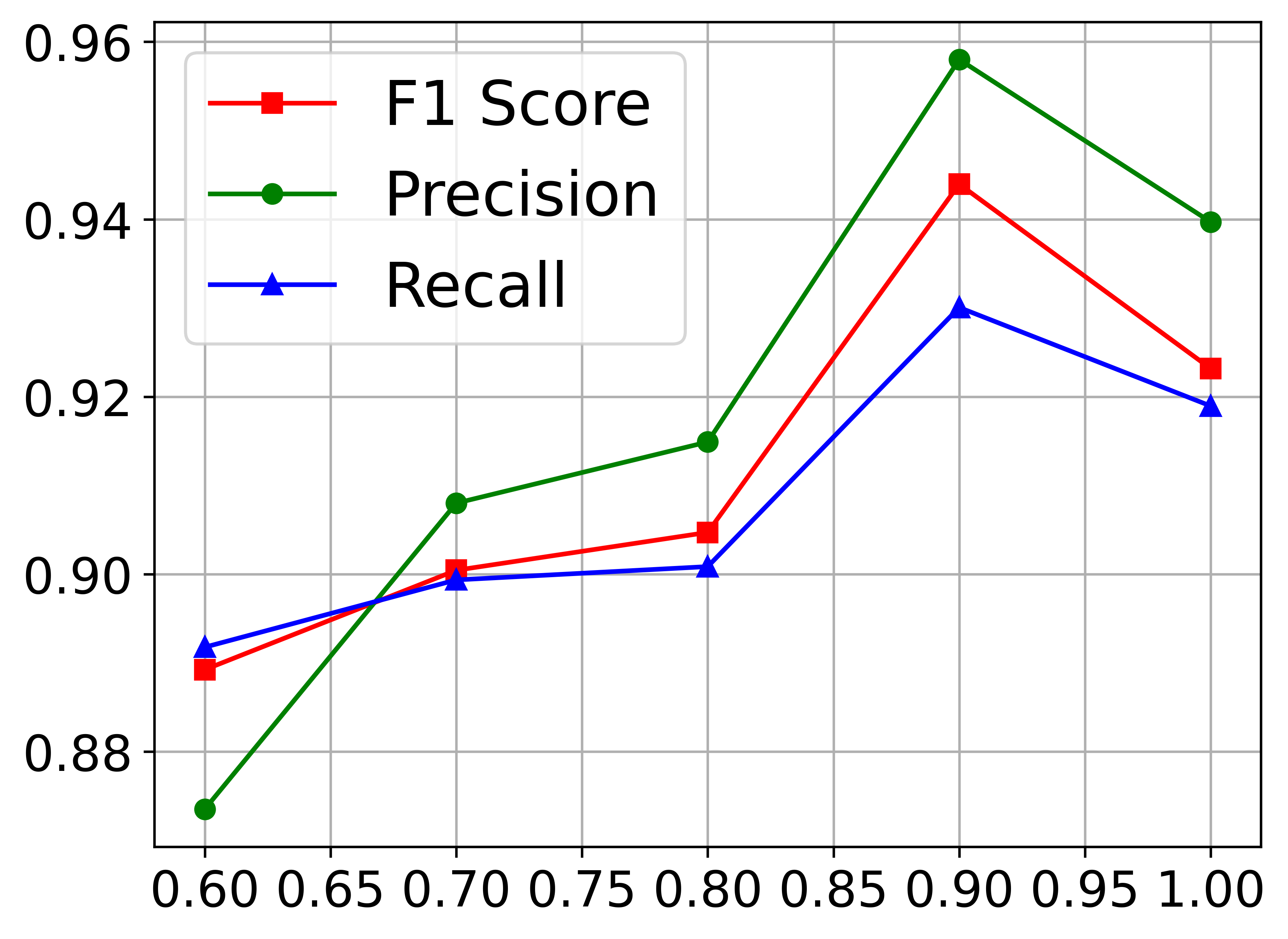}
  }
  \vspace{-2mm}
  \caption{The Sensitivity Analysis of Threshold $\alpha$ and $\beta$}
  \vspace{-4mm}
  \label{Sensitivity}
\end{figure*}

\subsubsection{\textbf{RQ4} The sensitivity of \nm to the parameters}

The threshold $\alpha$ is the parameter that determines the sparsity of the relational learning, while the threshold $\beta$, \textit{i.e.}, the parameter that determines the label of training data during positive unlabeled learning, may affect the performance by affecting the label distribution of training samples. We hereon evaluate the sensitivity of \nm to these two hyper-parameters on two industrial datasets. We change the value of $\alpha$ and $\beta$ while keeping all other parameters unchanged in our experiments to guarantee fairness. Specifically, we choose the value of $\alpha$ in an appropriate range from 0.3 to 0.7 at a step of 0.1. The value of $\beta$ is selected, ranging from 0.6 to 1 at a step of 0.1. When the value of $\beta$ is 1, it is equivalent to the variant without positive unlabeled learning since all samples will be seen as normal. When the value of $\beta$ is lower than 0.5, a large portion of samples will be labeled as anomalous and introduce severe noise to training data, which is not the case in the real scenario.

Figure {\ref{Sensitivity}} presents the experimental results of RQ4. For the threshold $\alpha$, the performance is relatively stable under different settings. It can be attributed to the distribution of learned graph attention scores, which are either close to 0 or 1. The polarization of graph attention scores means that the threshold chosen within the aforementioned range tends not to impact the performance in a significant manner, thus stable across different values of $\alpha$. This makes \nm easy to deploy in practice. For parameter $\beta$, a good threshold indeed helps improve the accuracy of our model. In dataset A, the best threshold is between 0.7 and 0.8, while in dataset B, the best threshold is between 0.8 and 0.9. This is because the anomaly ratio of Dataset B is slightly lower than Dataset A, so a lower amount of unlabeled anomalous samples exist in the training part. Compared to without positive unlabeled learning, a properly selected threshold would improve recall because some abnormal behaviors have been labeled as positive, and these abnormal patterns would be reported in the testing set, thus reducing false negatives. 

\subsubsection{\textbf{RQ5} The efficiency of \nm}

\begin{figure}
  \centering
  \subfigure[Training Time versus Metric Numbers]{
  \label{Training}
  \includegraphics[width=0.4\columnwidth]{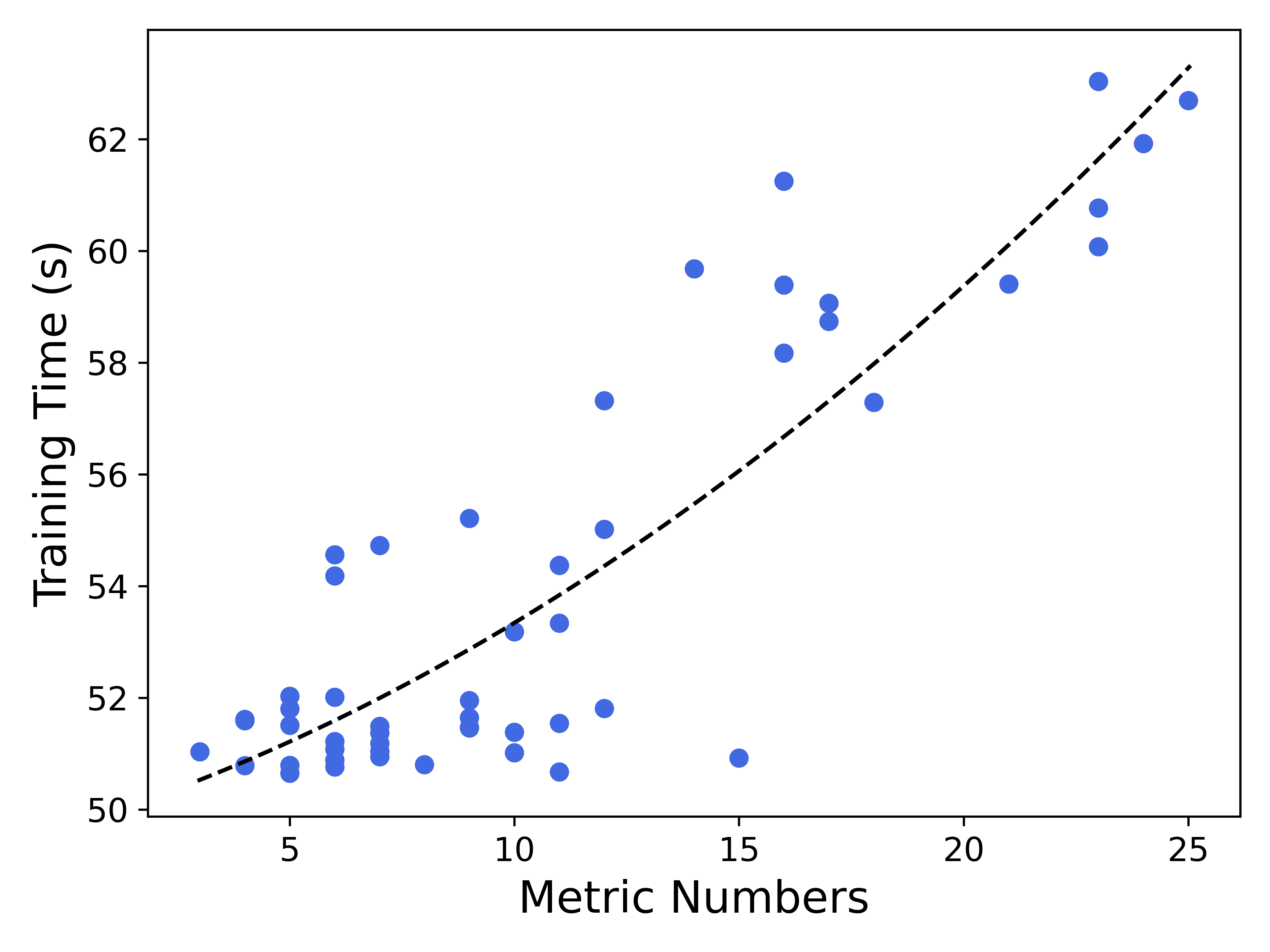}
  }
  \subfigure[Testing Time versus Metric Numbers]{
  \label{Testing}
  \includegraphics[width=0.4\columnwidth]{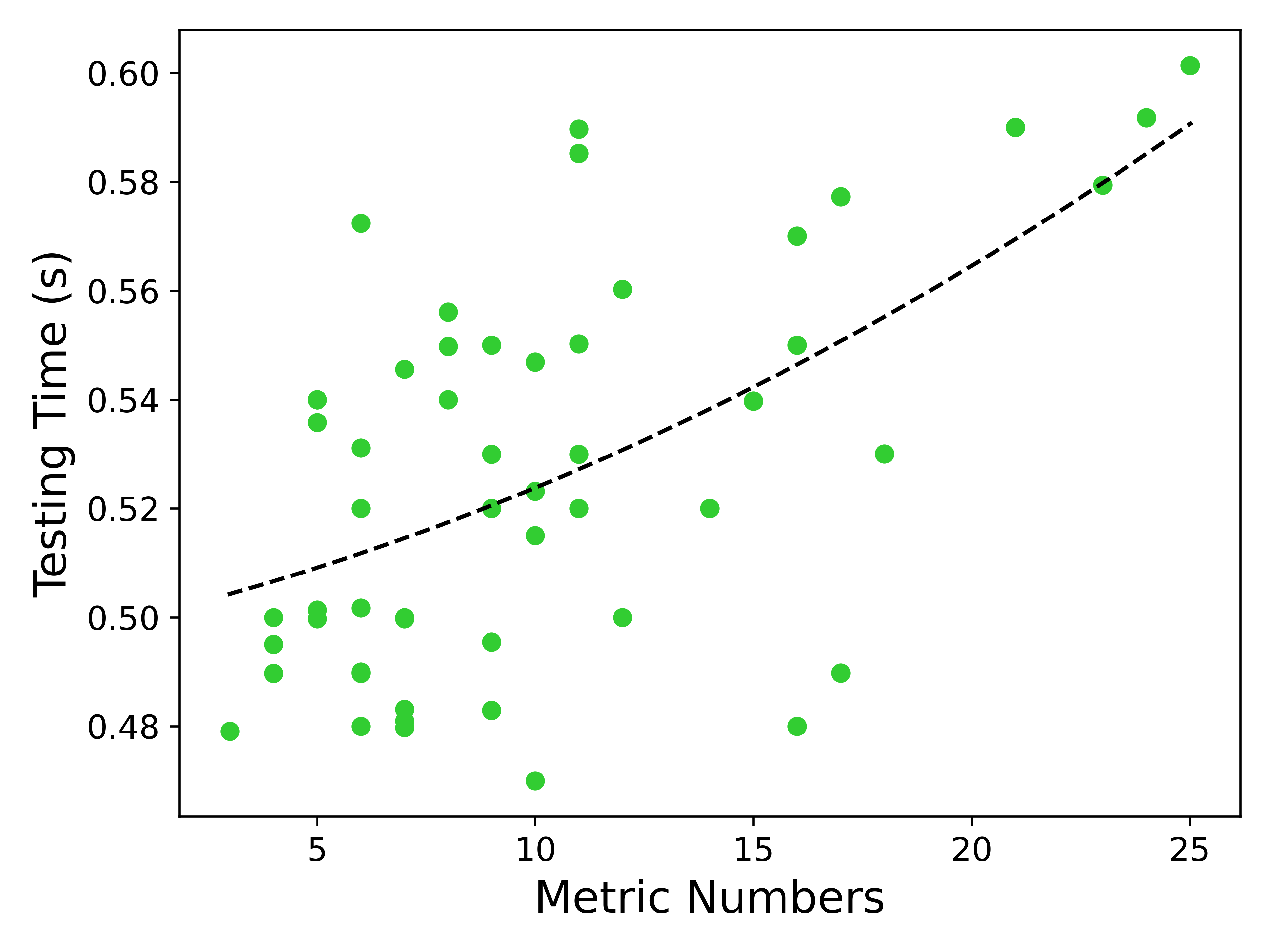}
  }
  \vspace{-2mm}
  \caption{The Efficiency Analysis of \nm}
  \vspace{-4mm}
  \label{Efficiency}
\end{figure}

In this section, we evaluate the efficiency of \nm regarding the number of metrics. We perform our method on the industrial dataset A and B and record the training and testing costs. The training time is defined as the total time of feeding the preprocessed data to the model, and the testing time is the total time used to predict whether there are performance anomalies on all the sliding windows in the test set. All the data samples are trained with the same batch size for 20 epochs. The correlation between training/testing time and metrics number is shown in Figure~\ref{Efficiency}, where we can observe that both of them are near quadratic correlations. In our scenario, the metrics collected for node-level and service-level performance anomaly detection are typically smaller than 100. According to the polynomial approximation, the training and testing time for data that contains 100 monitoring metrics is 192.9 seconds and 3.89 seconds, respectively. Typically, model retraining will not be triggered more frequently than once per day due to the computation cost. Thus, the training time of 192.9 seconds is affordable for industrial deployment. Furthermore, since the monitoring metrics are typically collected at an interval of 1 minute, the testing time of 3.89 seconds in the online detection phase is enough for real-time performance issues identification. It should be noted that all the experiments are run on a Linux server with only one Tesla V100 PCIe GPU, however, the efficiency is even further enhanced when utilizing the significant computational power that can be harnessed from thousands of GPUs available in cloud service systems.

\begin{figure}[t]
\centering
\includegraphics[width=.7\linewidth]{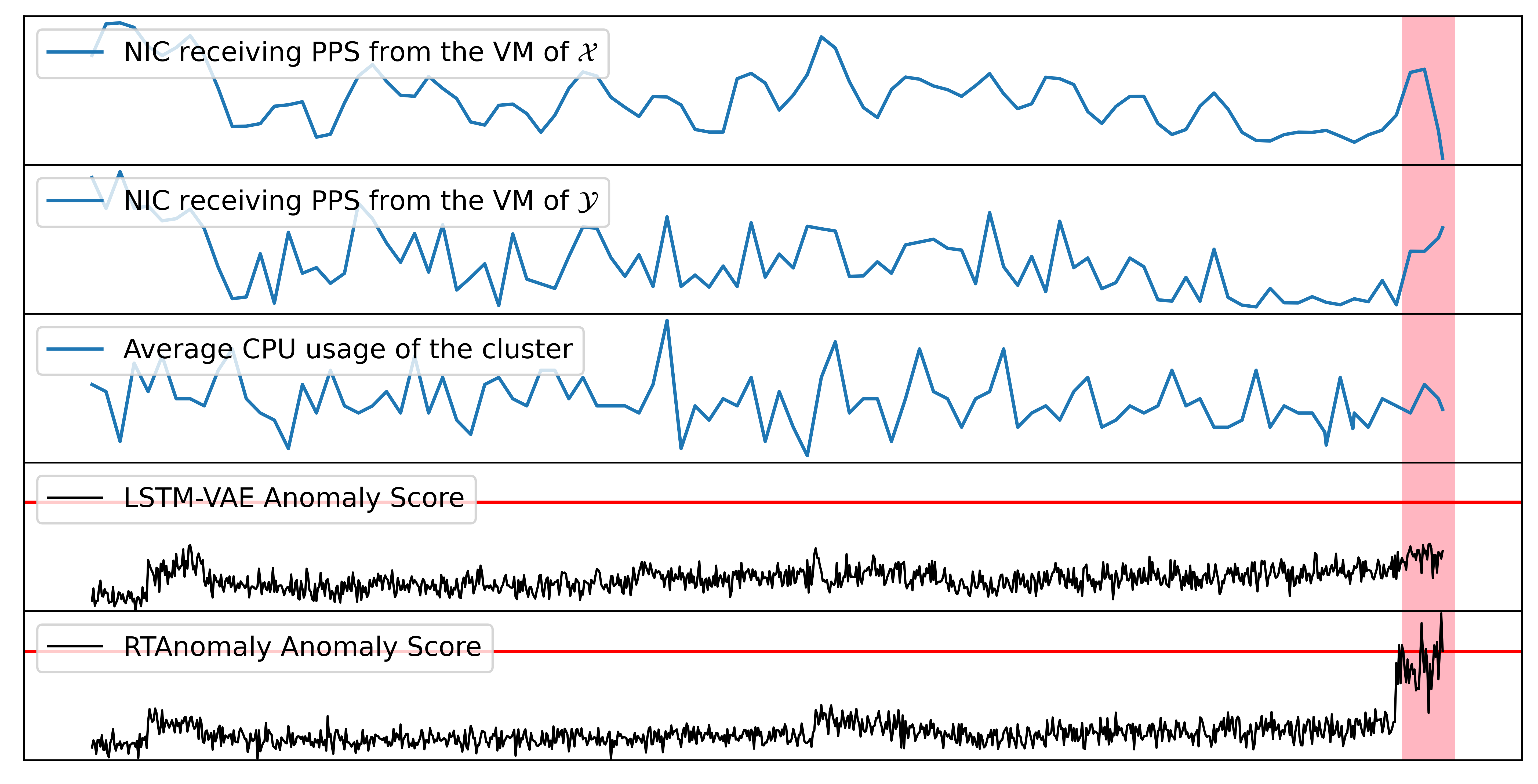}
\vspace{-2mm}
\caption{A case that \nm finds false negative}
\vspace{-2mm}
\label{Case Study 1}
\end{figure}

\begin{figure}[t]
\centering
\includegraphics[width=.7\linewidth]{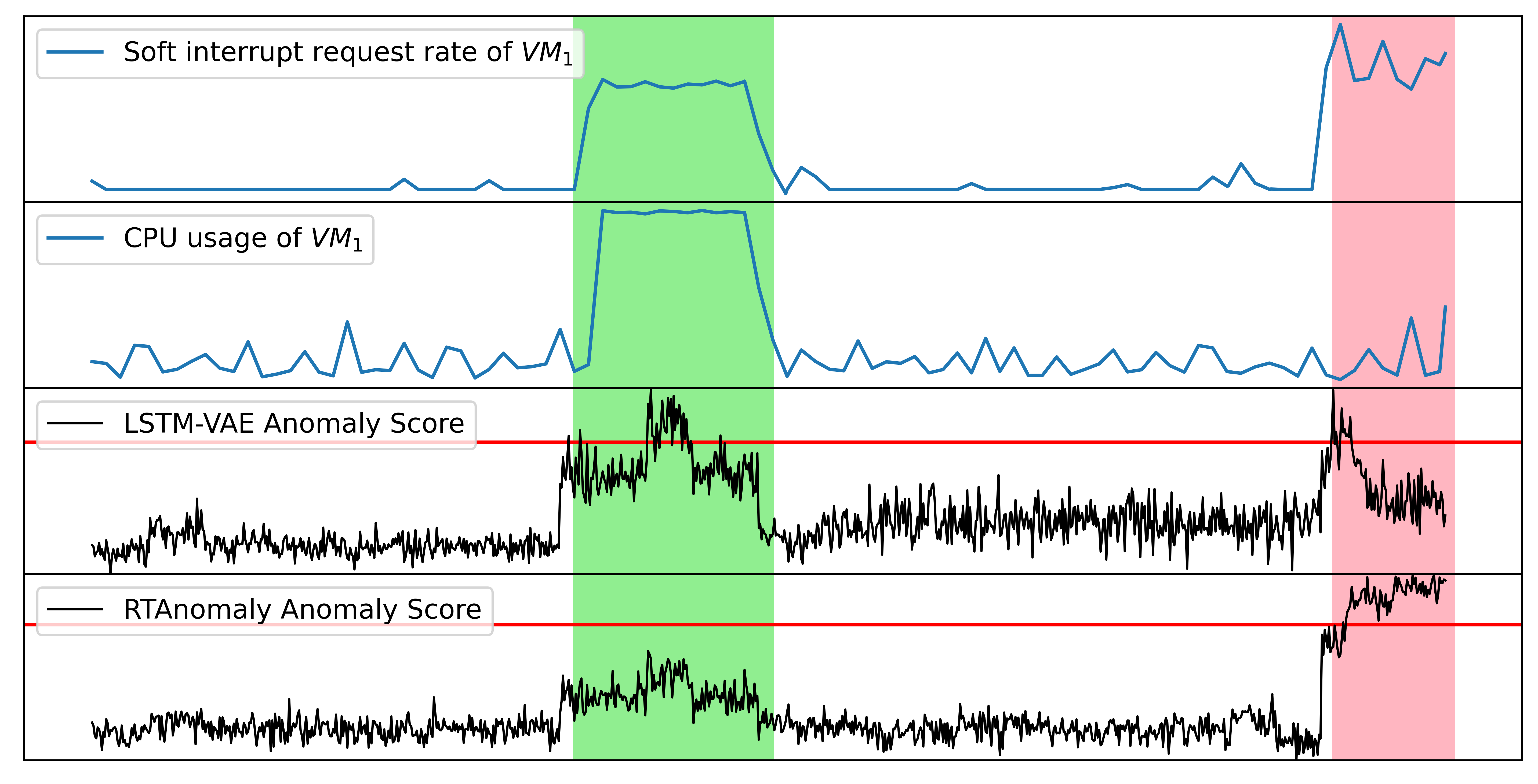}
\vspace{-2mm}
\caption{A Case that \nm avoids false positive}
\vspace{-4mm}
\label{Case Study 2}
\end{figure}

\subsection{Case Study}

To further demonstrate the effectiveness of \nm, we conduct a case study concerning two industrial cases on identifying performance anomalies and localizing the anomaly on metrics that violate their intrinsic correlation. The first case is shown in Figure~\ref{Case Study 1}, which is the same example in Section~\ref{background: correlation}. The first three rows show the monitoring metrics of the NIC receiving PPS from a virtual machine running microservice $\mathcal{X}$, the NIC receiving PPS from a virtual machine running microservice $\mathcal{Y}$, and the average CPU usage of all virtual machines of the ELB service. The last two rows show the anomaly score of the baseline method LSTM-VAE and our proposed \nm for comparison. This anomaly can hardly be discovered when observing each metric individually, as the correlation violation is critical to identifying this performance issue. However, our \nm can find the false negative that is neglected by baseline models. The first and second metrics can also be highlighted to provide engineers with further insight into the performance issue.

Another example is shown in Figure {\ref{Case Study 2}}. The first metric is the soft interrupt request rate of a virtual machine $VM_1$ running on an Elastic Cloud Server (ECS). The second metric is the CPU usage of $VM_1$. We can observe that the green segment seems like a performance issue, as the soft interrupt request rate of both $VM_1$ and the CPU usage all encounter spikes. However, this is usually caused by increased user requests and should not be regarded as a performance issue. In contrast, since the correlations between these metrics are not violated, an anomaly alert should not be triggered. The red segment represents a soft interrupt issue because the CPU usage of $VM_1$ is not synchronized with the increase of the soft interrupt request rate of $VM_1$. Compared with the baseline method LSTM-VAE, \nm can produce a significantly higher anomaly score in the true anomaly labeled with the red area, thus being more effective. In this case, though both \nm and baseline methods like LSTM-VAE can detect the true anomaly, our \nm can avoid sending a false alarm to engineers. This is helpful for the engineers to mend the service system because it prevents unnecessary interventions.

\section{DISCUSSION}

In this section, we share the success story of our deployment of \nm in the industrial environment of the service system of Huawei Cloud and discuss the threats and limitations of our approach. 

\subsection{Industrial Experience}

In this section, we share our experience in applying \nm to the real-world cloud system of Huawei Cloud, a full-stack cloud system that consists of an infrastructure layer, a platform layer, and an application layer, aiming to demonstrate the practical usefulness of \nm. Huawei Cloud serves hundreds of millions of cloud service tenants, offering them low-latency and high-performance services that span computing, storage, networking, and database solutions. Among them, ELB is a crucial service that is tasked with the automatic distribution of incoming network traffic across a multitude of targets, including Elastic Cloud Server (ECS) instances, containers, and IP addresses across different Availability Zones (AZs). Relational Database Service (RDS) is another reliable and scalable managed DB service that frees up developers from handling time-consuming database administration tasks. In addition, the API Gateway Service (APIG), is an API management service that serves as a single point of entry into cloud systems, sitting between the user and a collection of backend services. It receives requests from an application user, routes the request to the appropriate services, gathers the appropriate data, and combines the results for the user in a single package. The Object Storage Service (OBS) provides stable, secure, efficient, and easy-to-use object storage. It enables storage and retrieval of any amount of data at any time, facilitating cloud storage for applications, big data analysis, and backup and archiving scenarios. The reliability of these services is among the most crucial concerns for Huawei Cloud and its tenants.

\begin{table}[t]
\centering
\vspace{-8pt}
\caption{A Summary of the Node Types in the Huawei Cloud}
\scalebox{0.85}{\begin{tabular}{m{0.2\columnwidth}<{\centering}|m{0.75\columnwidth}<{\centering}}
\hline
Node Type & Description \\  
\hline
Management Nodes & Management nodes are used to deploy FusionSphere OpenStack, an enterprise-level platform to enhance computing, storage, network management, installation and maintenance, security, and reliability while supporting multiple infrastructure virtualization technologies at the resource pool layer. Management nodes use UVP as the host OS. The Computing cloud services, storage cloud services, network cloud services, common components, and management domain components are deployed on VMs.\\
\hline
Network Nodes & The network node uses the UVP as the host OS. The virtual Router, L3NAT, L3 service, and VPN components are deployed on VMs.\\
\hline
Compute Nodes & Compute nodes can be divided into two subtypes. The first is the KVM compute node for general-purpose Elastic Computing Services (ECS) in tenant VMs. The second is the KVM compute node for GPU-accelerated ECSs which provision GPUs for deep learning jobs in tenant VMs. These two types of KVM compute nodes use the UVP as the host OS, and FusionSphere OpenStack (role compute) is also deployed.\\
\hline
Storage Nodes & Distributed storage nodes in Huawei typically support Elastic Volume Service (EVS). After the deployment of Huawei Distributed Block Storage, this node is utilized by the EVS service to provision EVS instances in tenant EVS disks.\\
\hline
\end{tabular}}
\vspace{-8pt}
\label{Node Type}
\end{table}

\begin{figure}[t]
\centering
\includegraphics[width=.7\linewidth]{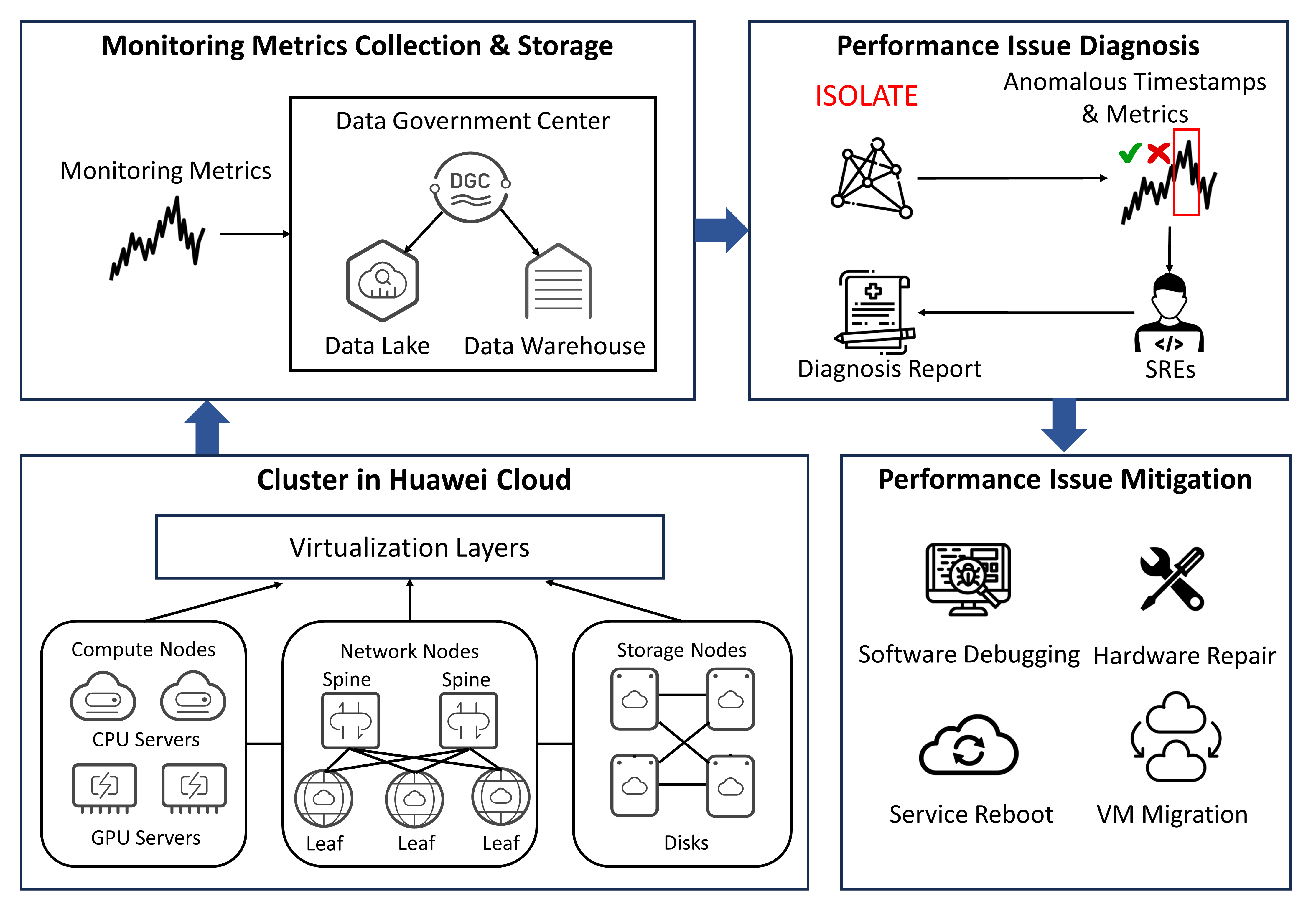}
\vspace{-2mm}
\caption{The Pipeline of Deploying \nm in Huawei Cloud System}
\vspace{-6mm}
\label{Industrial}
\end{figure}

\nm has been successfully incorporated into the performance issue detection system of large-scale online service systems in Huawei Cloud since \textit{November 2022}, specifically, the overall pipeline of deployment is shown in Figure~\ref{Industrial}. The clusters in Huawei Cloud represent a collection of nodes that work together to provide various services, ensuring high availability, reliability, and scalability. Particularly, there are four types of nodes in Huawei Cloud: Management node, network node, compute node, and storage node, where the details of them are elaborated in Table~\ref{Node Type}. For clusters in cloud service systems, it has been a common practice for monitoring metrics used to profile the runtime status~\cite{ren2019time, islam2021anomaly, yan2022cmmd}. Thus, software reliability engineers usually collect tens of monitoring metrics (like CPU usage, network traffic, disk I/O, request rates, and memory usage) in nodes of the cluster through monitoring tools like Grafana, Prometheus, etc~\cite{agarwal2023outage}. Then, these monitoring metrics are stored in the Data Lake of Huawei Cloud, a highly scalable and flexible storage system that consists of the Data Lake Storage, the Data Warehouse, and the Data Lake Governance Center (DGC). Data Lake Storage is the actual storage space where all the data, including the monitoring metrics, are stored while the Data Warehouse is an enterprise system used for reporting and data analysis. On top of these two components, the DGC is responsible for managing the data stored in the data lake that oversees the lifecycle of the data, from ingestion and storage to usage and deletion. With Huawei Cloud's data lake, real-time data analysis is enabled, \ie as soon as monitoring metrics are collected and stored in the data lake, they can be immediately accessed and analyzed by the performance issue diagnosis system empowered with \nm. The results of \nm provide not only the timing of a performance issue but also the metrics that are most related to this issue. Then, alerts will be triggered immediately and sent to the SREs for further investigation. The SREs first inspect the alert and the associated metrics to understand the mechanism of the problem and find the root cause. Once the root cause has been identified, the SREs prepare a diagnosis report, including a detailed description of the performance issue, the associated metrics, the identified root cause, and potential mitigation strategies. For instance, if the performance issue is a memory leak, engineers may need to perform software debugging involving tracing the program's execution and implementing appropriate code fixes to prevent the leak from happening again. As another example, if the problem is network congestion, the mitigation strategies may include hardware repair like replacement of the NICs.

\begin{figure}[t]
\centering
\includegraphics[width=.6\linewidth]{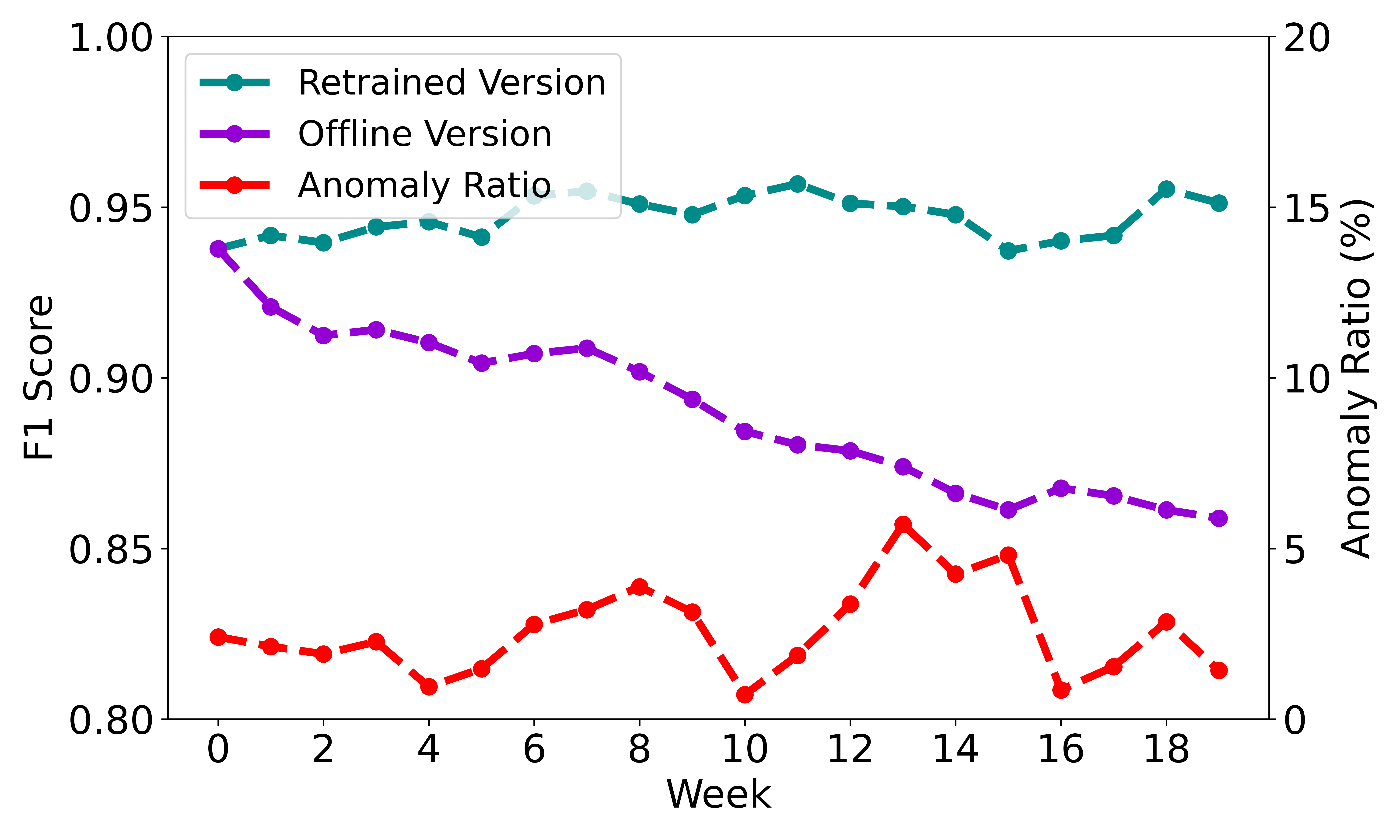}
\vspace{-2mm}
\caption{The Performance of deploying \nm in Huawei Cloud System}
\vspace{-6mm}
\label{Deployment}
\end{figure}

Due to frequent service updates and changes in user behavior, monitoring metric patterns may also evolve, a phenomenon known as concept drift. This drift can gradually weaken the effectiveness of \nm over time. To alleviate this issue, \nm is retrained on a weekly basis with newly collected data. Figure~\ref{Deployment} illustrates the performance issue identification accuracy of both the retrained and offline versions over 20 weeks. A noticeable downward trend in the accuracy of the offline version can be observed, attributed to pattern drift. Conversely, the version that is periodically retrained displays relatively stable performance. This adaptability to new patterns in online deployment scenarios underscores the robustness and effectiveness of \nm within dynamically evolving cloud systems. More specifically, take one of the services $\mathcal{R}$ for example, the time proportion of undergoing performance issues during this period is 3.4\%, and the online version of \nm achieves a precision of 0.904, recall of 0.951, and F1 of 0.926. Another interesting observation is that though the anomaly ratio analyzed by engineers ranges from 0.7\% to 5.6\% during the 20-week period, the performance remains relatively stable in the retrained version of \nm, demonstrating that the performance of \nm is not sensitive to the anomaly ratio. It should be noted that the weekly retraining enhances the adaptiveness of our method, with an additional cost of human labeling of the potential performance anomaly data. However, it has been a common practice in cloud systems that on-call SREs manually verify some reported suspicious performance anomaly~\cite{chen2020towards, wang2021fast, liu2023scalable}. Specifically, according to the interview with SREs, they typically spend approximately one hour per week checking these data and they consider it to be affordable.

\begin{figure}[t]
\centering
\includegraphics[width=.6\linewidth]{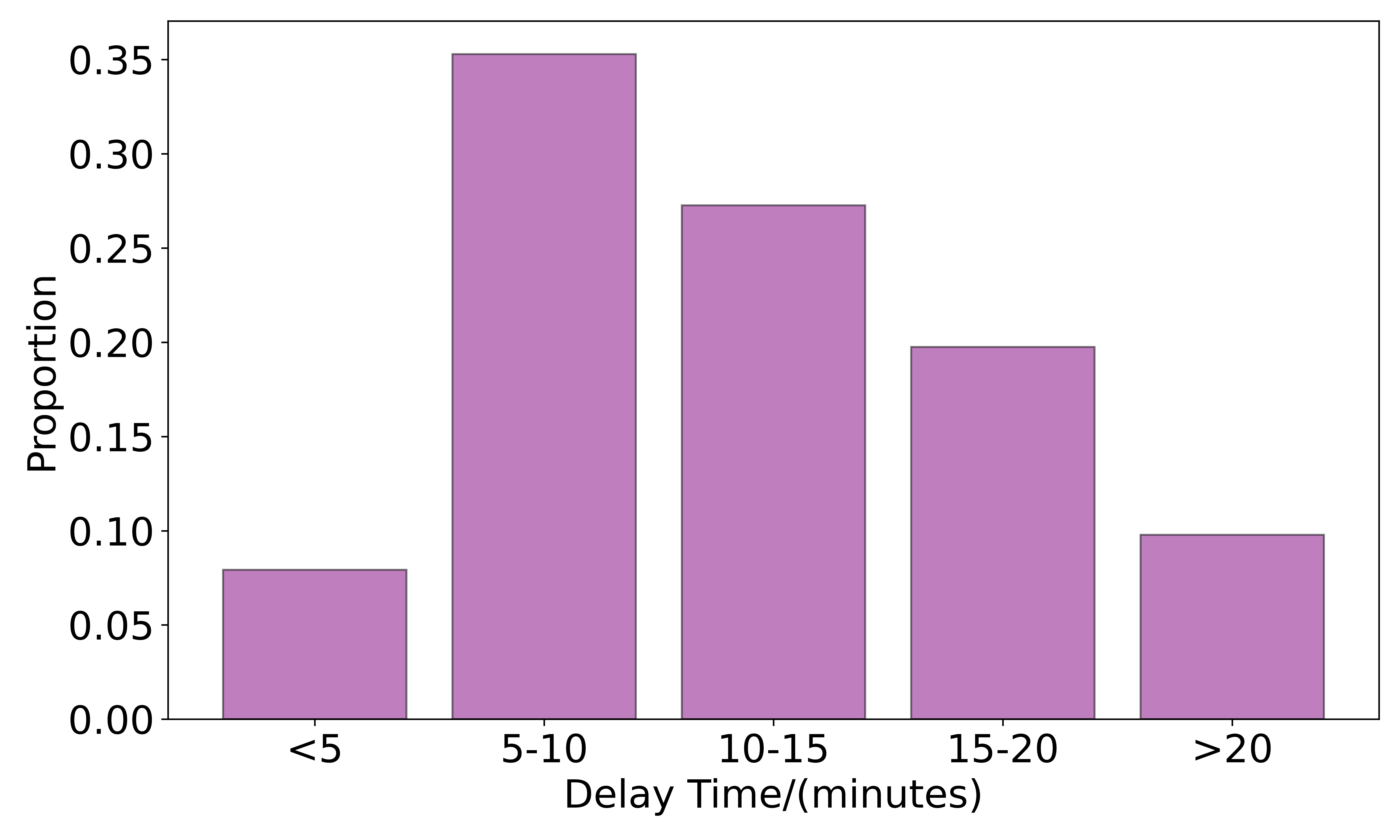}
\vspace{-2mm}
\caption{The Delay Time of \nm in Identifying Performance Issues}
\vspace{-6mm}
\label{Delay Time}
\end{figure}

The ability to timely alert operation engineers after a performance issue happens is also of great significance. The delay time of \nm in Huawei Cloud is the time delay between the detection of performance issues and the happening time of the issues confirmed by customer tickets~~\cite{liu2024mtad}. The delay time of \nm is shown in Figure~\ref{Delay Time}, where we can observe that most of the performance issues can be captured 5 to 20 minutes after the happening and SREs can take mitigation strategies to avoid continuously compromising the user experience. Consequently, the potential negative impacts on the QoS can be significantly mitigated, enhancing system reliability and ensuring high availability.

\subsection{Limitations of ISOLATE}

Our approach \nm employs graph attention to extract the correlation between metrics and the PU Learning strategy to identify positive samples in unlabeled training data. Though satisfactory performance in detecting performance anomalies is achieved in our experiments, we have identified several limitations of our approach and the challenges that arise in real-world applications.
 
Firstly, our approach is not adaptive to the evolving metric patterns and correlation in cloud systems due to rapid software and application upgrades~\cite{chen2022adaptive}. Thus, our model can result in false positives when new patterns or correlations are not included in the training dataset. Fortunately, we found that regular model retraining, \eg retraining every week, works well to learn the new behaviors of the cloud system and thus overcome this limitation~\cite{harsh2023murphy}.

Secondly, the complexity of large-scale cloud systems results in a vast number of monitoring metrics, which escalates the computational costs of \nm. Fortunately, our approach is deployed at the node level or service level, where the number of metrics is typically fewer than 100. This reduces computational demands and enhances efficiency. It's important to note that performance issues detected at the node or service level can serve as indicators of system-level performance anomalies, offering valuable insights and early warnings about potential system failures, which is useful for SREs.

Thirdly, our approach lacks interpretability in traditional methods such as decision trees, which can be a limitation to some content. However, it's worth noting that our method is not entirely without interpretability. The correlation violations between metrics can provide engineers with more granular information about anomalous metrics, offering some level of insight into the root cause localization of performance issues, and making it partially interpretable.

\subsection{Threats to Validity}

\textbf{Internal threats.} The correctness of the implementation of baselines constitutes one of the internal threats to our study's validity. For the baselines, we utilized the open-sourced code released by the authors of the papers or packages on GitHub like {\cite{zhao2019pyod}}. As for our proposed approach, the source code has been reviewed meticulously by the authors, as well as several experienced software engineers, to minimize the risk of errors and increase the overall confidence in our results. For parameter selection, we conducted extensive experiments with different parameters to find the most suitable configurations for both baselines and our proposed method. We chose the parameters based on the best results obtained in these experiments. To make our results reproducible, we have also made our code and partial data available.

\noindent\textbf{External threats.} The external threats to the validity of our study mainly lie in the generalizability of our experimental results. We conduct experiments on the large-scale online systems of two regions within a prominent cloud service company. In addition to this, our approach is also evaluated on a publicly available dataset containing monitoring metrics from an Internet company, further expanding the scope of our evaluation. While the diversity of the experimental settings provides some confidence in the generality of our findings, it is essential to acknowledge that results might vary when applied to different cloud service providers, industries, or specific use cases. Nevertheless, we believe that our experimental results, obtained from these multiple sources, can demonstrate the generality and effectiveness of our proposed approach, \nm.

\section{RELATED WORK}

Detecting performance issues on monitoring metrics for online service systems has been a hot topic. Monitoring metrics used to reflect the run-time status of the whole system are usually denoted as multivariate time series. The main challenge of multivariate metrics anomaly detection is twofold: First, effective modeling of complex relational and temporal dependency~\cite{wu2023understanding}. Second, noise data will be introduced inevitably during the manual labeling process. It is hard to get rid of these noises. Related studies can be categorized into machine learning or signal processing-based and deep learning-based approaches.

\textit{Machine Learning/Signal Processing Methods.} OCSVM {\cite{scholkopf2001estimating}} is a clustering-based method that learns the boundary for the normal data without anomalous samples and identifies the data outside the border as anomalies. Isolation Forest (iForest) {\cite{liu2008isolation}} applies multiple isolation trees and ensembles them based on the assumption that anomalies should be rare and isolated from normal observations with very short heights. Local Outlier Factor (LOF) is a density estimation-based anomaly detection approach that calculates the local density. The samples with an extremely lower density compared to their neighbors would be recognized as anomalies. JumpStarter~\cite{ma2021jump} is a signal processing-based method that adopts the compressed sensing technique to reconstruct the input data. It adopts a shape-based clustering strategy to reduce the volume of data and utilizes an outlier-resistant sampling to avoid sampling anomalous values.

\textit{Deep Learning Methods.} With the surge of deep learning~\cite{zhang2024curvature, zhang2023transferable}, there has been a variety of studies in applying deep learning to conduct anomaly detection on multivariate metrics data. A deep autoencoder with a Gaussian mixture model (DAGMM) {\cite{zong2018deep}} is utilized to detect anomalous data points. MSCRED~\cite{zhang2019deep} proposes a multi-scale convolutional recurrent encoder-decoder that detects anomalies. The reconstruction error of the input time series is utilized to diagnose anomalies. To detect performance degradation anomalies in software systems, Long Short-Term Memory (LSTM) has been deployed to guarantee high performance in {\cite{zhao2021predicting}}. LSTM-VAE {\cite{park2018multimodal}} combines the LSTM networks and the VAE to reconstruct the distribution of sliding windows from multivariate metrics regardless of the temporal dependencies in time series {\cite{chen2022adaptive}}. Similarly, LSTM-NDT {\cite{hundman2018detecting}} leverages Long Short-Term Memory (LSTM) networks with non-parametric dynamic thresholds to pursue the reliability of the systems. OmniAnomaly {\cite{su2019robust}} extends the LSTM-VAE with a normalizing flow and utilizes the reconstruction error for detection. However, the capability of this approach is degraded when there is severe noise in the training metric. THOC {\cite{shen2020timeseries}} is a model that fuses the temporal features at multi-scale from intermediate layers by hierarchical clustering and detects the anomalies by the multi-layer distance loss. MTSAD {\cite{wang2022active}} is another model combining active learning with existing VAE-based anomaly detection, which detects the difference between normal and abnormal samples in reconstruction error and latent space. TranAD~\cite{tuli2022tranad} is a transformer-based anomaly detection model that adopts attention-based sequence encoders to swiftly perform inference with the knowledge of the temporal trends in the data. It further leverages self-conditioning and adversarial training to gain training stability. ACVAE~\cite{zhang2024acvae} is a variational autoencoder-based method that combines adversarial mechanisms and contrast learning. The former constrains the decoder and gains some discriminatory power, while the latter allows the encoder to obtain more training samples. SLA-VAE~\cite{huang2022semi} is a semi-supervised VAE-based model that employs convolutions to capture inter-metric dependence and utilizes active learning to update the VAE model using a small number of uncertain samples.

Recently, among the deep learning-based anomaly detection model, Graph Neural Networks (GNNs) based methods have shown promising potential to effectively capture temporal and spatial dependencies among metric pairs of multivariate metrics~\cite{jin2023survey}. MTAD-GAT~\cite{zhao2020multivariate} is the first work that utilizes the graph-attention network to model the relationships between sensors in addition to jointly optimized reconstruction and forecasting-based models. Another representative model is the Graph deviation network (GDN)~\cite{deng2021graph}, which learns the pairwise relationship through cosine similarity and models the time series as a graph through an adjacent matrix. It predicts future values and computes the forecast error as an anomaly score. GTA~\cite{chen2021learning} is a Transformer-based framework for anomaly detection that automatically learns sensor dependencies. The authors propose an Influence Propagation (IP) graph convolution and multi-branch attention technique that improves the training efficiency. FuSAGNet~\cite{han2022learning} utilizes a sparse autoencoder to extract latent feature representations. It then predicts future time series from the sparse representations and graph structures learned from graph neural networks. TopoMAD~\cite{he2020spatiotemporal} is a topological-aware model that integrates graph neural networks, LSTM and VAE. The topological information is extracted through the pod-based division in a Kubernetes microservice system or pods that share similar behaviors. It should be noted that the topology in Huawei Cloud is highly dynamic due to frequent deployment changes and software updates, which makes it challenging to determine. Thus, TopoMAD does not apply to our scenario.

Furthermore, many efforts have been devoted to multi-source data-based anomaly detection in microservice systems~\cite{lee2023heterogeneous}. For example, SCWarn~\cite{zhao2021identifying} is proposed to identify bad software changes in online service systems via multimodal anomaly detection from metrics and logs. It serializes the metrics and logs separately and extracts the temporal dependency through the LSTM model. AnoFusion~\cite{zhao2023robust} is another unsupervised failure detection approach for microservice systems. It employs a Graph Transformer Network (GTN) to capture correlations within the heterogeneous multimodal data. It then seamlessly integrates a Graph Attention Network (GAT) with a Gated Recurrent Unit (GRU) to model the temporal information of dynamically changing multimodal data. Though three types of monitoring data (metrics, logs, and traces) can be utilized to ensure the reliability of microservice systems~\cite{picoreti2018multilevel}, the real-time collection of these three types of data can be challenging.

Though some GNN-based methods like~\cite{deng2021graph, zhao2020multivariate} capture both the temporal dependencies and relational dependencies, performance issues can be indicated by the violation of correlation in our scenario, which is not taken into account explicitly in existing approaches. Besides, the performance of these models will degrade due to the existence of unlabeled noise in training data, where noise is prevalent in our scenario.

\section{CONCLUSION}

In this work, we propose \nm, a novel framework to mine correlations among metrics, detect performance issues, and localize the correlation-violation metrics. Specifically, \nm leverages a graph neural network with graph attention to capture the complex correlations between a variety of metrics and a label-conditional VAE model to distinguish normal and abnormal patterns. We also propose to utilize the positive unlabeled learning strategy to overcome the impacts of noisy data. Extensive experiments on one public dataset and two industrial datasets show that \nm achieves 0.945 F1-Score on anomaly detection and 0.920 Hit@3 in terms of localizing correlation-violation metrics, outperforming all the baselines. Furthermore, our framework has been successfully incorporated into Huawei Cloud’s performance issue monitoring and detection system. Both the codes and data are released to facilitate future research.

\section*{ACKNOWLEDGMENTS}

The work described in this paper was supported by the Research Grants Council of the Hong Kong Special Administrative Region, China (No. CUHK 14206921 of the General Research Fund) and Fundamental Research Funds for the Central Universities, Sun Yat-sen University (No. 76250-31610005).

\bibliographystyle{ACM-Reference-Format}
\bibliography{tosem24}

\end{document}